\documentclass[conference]{IEEEtran}

\usepackage{acro} 
\usepackage{times}

\usepackage{multicol}
\usepackage[bookmarks=true]{hyperref}
\usepackage{times}

\usepackage[numbers,sort&compress]{natbib}
\usepackage{multicol}
\usepackage[bookmarks=true]{hyperref}
\usepackage{gensymb}
\usepackage{float}
\usepackage{bm}
\usepackage{gensymb}
\usepackage[binary-units,product-units=power]{siunitx}
\usepackage{graphicx}
\usepackage{mathtools}
\usepackage{amsmath}
\usepackage{amssymb}
\usepackage{amsfonts}
\usepackage[mode=buildnew]{standalone}
\usepackage{booktabs} 
\usepackage[caption=false,font=footnotesize]{subfig}
\usepackage{nccmath}
\usepackage{algorithm}
\usepackage{algorithmic}

\usepackage{cleveref}
\usepackage{lipsum}
\usepackage{capt-of}

\DeclareAcronym{DOF}{
  short = DOF,
  long  = degrees of freedom
}
\DeclareAcronym{GNSS}{
  short = GNSS,
  long  = global navigation satellite system
}

\DeclareAcronym{IMU}{
  short = IMU,
  long  = Inertial Measurement Unit
}
\DeclareAcronym{LiDAR}{
  short = LiDAR,
  long  = Light Detection and Ranging
}
\DeclareAcronym{UXO}{
  short = UXO,
  long  = unexploded ordnance
}

\DeclareAcronym{EKF}{
  short = EKF,
  long  = Extended Kalman Filter
}
\DeclareAcronym{iEKF}{
  short = iEKF,
  long  = iterated Extended Kalman Filter
}

\DeclareAcronym{LIO}{
  short = LIO,
  long  = LiDAR-Inertial Odometry
}

\DeclareAcronym{LO}{
  short = LO,
  long  = LiDAR Odometry
}

\DeclareAcronym{MAV}{
  short = MAV,
  long  = micro aerial vehicle,
  short-indefinite  = an
}

\DeclareAcronym{FOV}{
  short = FOV,
  long  = field of view
}

\DeclareAcronym{SDF}{
  short = SDF,
  long  = signed distance field,
  short-indefinite  = an
}

\DeclareAcronym{NCCR}{
  short = NCCR,
  long  = National Center of Competence in Research,
  short-indefinite = an,
  long-indefinite = a
}

\DeclareAcronym{SNR}{
  short = SNR,
  long  = Signal-to-Noise Ratio,
}

\DeclareAcronym{MID}{
  short = MID,
  long  = Mean Image Distance,
}

\begin{document}

\IEEEoverridecommandlockouts 

\title{BIEVR-LIO: Robust LiDAR-Inertial Odometry through Bump-Image-Enhanced Voxel Maps
\vspace{-0.25cm}}
\author{
\IEEEauthorblockN{Patrick Pfreundschuh$^{1}$, Turcan Tuna$^{2}$, Cedric Le Gentil$^{3}$, Roland Siegwart$^{1}$, Cesar Cadena$^{2}$, Helen Oleynikova$^{1,3}$}
\IEEEauthorblockA{$^{1}$Autonomous Systems Lab, ETH Z\"urich, Switzerland, $^{2}$Robotic Systems Lab, ETH Z\"urich, Switzerland, \\ $^{3}$Mobile Robotics Lab, ETH Z\"urich, Switzerland \vspace{-1.5mm}}

\thanks{This work was supported in part by the Swiss National Science Foundation (SNSF) NCCR DFab P3.\vspace{-0.5mm}}
\vspace{-0.75cm}}
\maketitle
\begin{abstract}
Reliable odometry is essential for mobile robots as they increasingly enter more challenging environments, which often contain little information to constrain point cloud registration, resulting in degraded LiDAR–Inertial Odometry (LIO) accuracy or even divergence.
To address this, we present BIEVR-LIO, a novel approach designed specifically to exploit subtle variations in the available geometry for improved robustness. 
We propose a high-resolution map representation that stores surfaces as voxel-wise oriented height images. This representation can directly be used for registration without the calculation of intermediate geometric primitives while still supporting efficient updates. 
Since informative geometry is often sparsely distributed in the environment, we further propose a map-informed point sampling strategy to focus registration on geometrically informative regions, improving robustness in uninformative environments while reducing computational cost compared to global high-resolution sampling.
Experiments across multiple sensors, platforms, and environments demonstrate state-of-the-art performance in well-constrained scenes and substantial improvements in challenging scenarios where baseline methods diverge. Additionally, we demonstrate that the fine-grained geometry captured by BIEVR-LIO can be used for downstream tasks such as elevation mapping for robot locomotion.

\end{abstract}

\IEEEpeerreviewmaketitle

\vspace{1mm}
\section{Introduction}
\label{sec:introduction}
\vspace{-1mm}
Ego-motion estimation is a fundamental building block of many mobile robotics applications. LiDAR–Inertial Odometry (LIO) has become one of the most widely adopted solutions due to its high accuracy and robustness in both indoor and outdoor environments \cite{ebadi2023present, lee2024lidar}. With the advent of smaller and more affordable 3D LiDAR sensors, LIO systems are no longer confined to research platforms but are increasingly deployed in consumer products, such as robotic lawnmowers.
Despite its success, the robustness of LIO degrades significantly in geometrically less informative environments, such as straight tunnels or large flat terrains, where the observed geometry does not sufficiently constrain point cloud registration. Several works have addressed this issue by detecting degeneracy \cite{nubert2022learning, tuna2023x}, but rely on auxiliary odometry from additional sensors, increasing system complexity. Recent methods attempt to mitigate degeneracy using LiDAR-inertial data alone~\cite{huang2024lio,lee2025lodestar} but only address short sections of degeneracy.
Another line of research leverages LiDAR intensity as an additional cue for registration \cite{khedekar2025pg, zhang2023ri, pfreundschuh2024coin}. While such intensity-augmented methods achieve promising performance, they depend on dense LiDAR data and are therefore incompatible with sensors that employ sparse or irregular scan patterns.

\begin{figure}[t]
    \vspace{3mm}
    \includegraphics[width=\linewidth, trim={0 12mm 0 2mm}, clip]{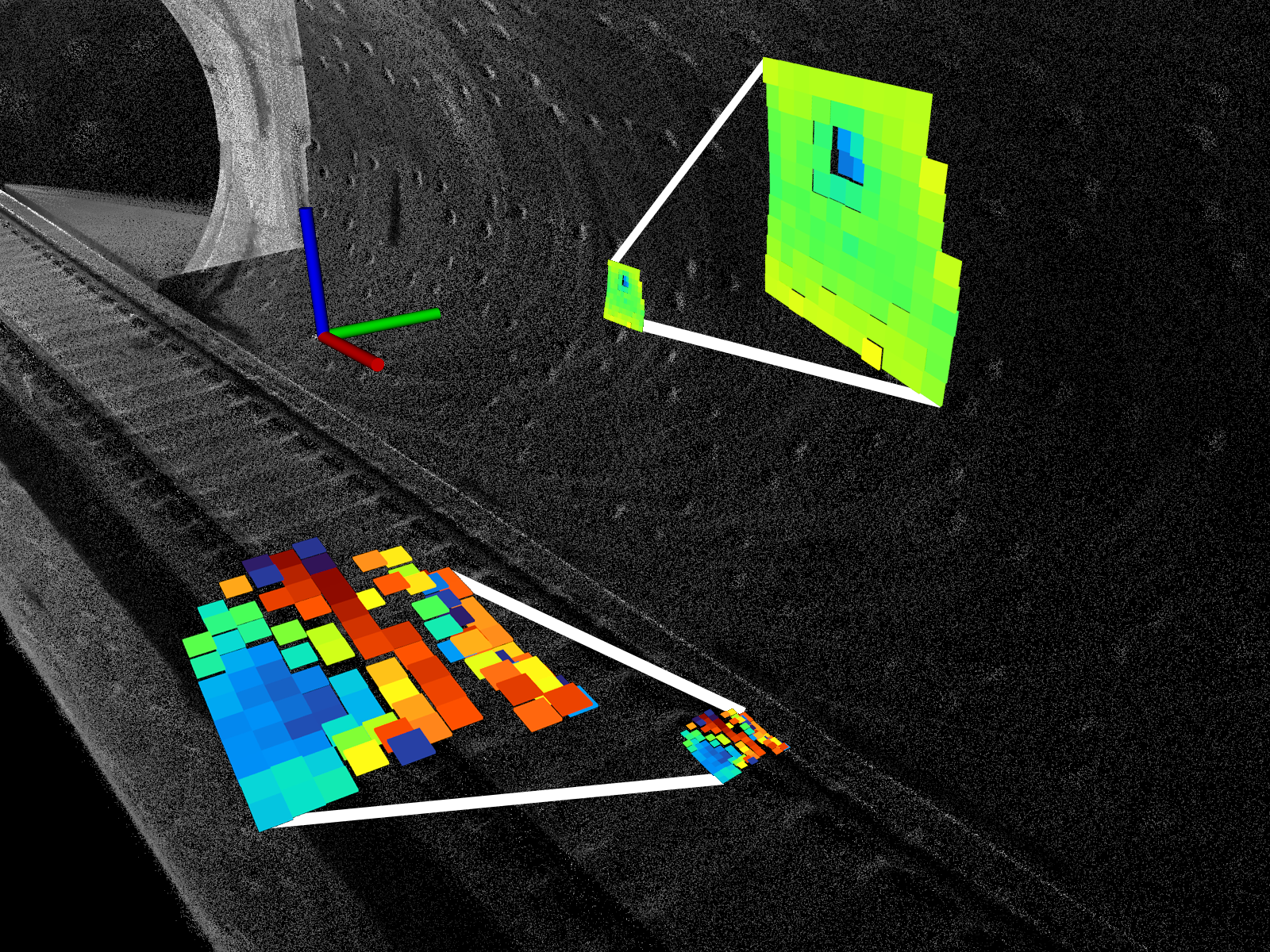}
    \vspace{-7mm}
    \caption{BIEVR-LIO achieves robust odometry in a tunnel through registration against its high-resolution bump-image–enhanced voxel map. The colorized bump images illustrate geometric details of wall cutouts and the train bed.}
    \vspace{-2mm}
    \label{fig:highlight}
    \vspace{-3mm}
\end{figure}

In this work, we approach robust LIO in geometrically sparse environments from a different perspective. We observe environments often considered uninformative are rarely degenerate in a strict sense, as real-world scenes typically contain subtle geometric variations that render the problem well-constrained in principle, as shown in \Cref{fig:highlight}. However, existing LIO systems fail to exploit this information because their map representations lack the resolution required to capture such fine-grained structure under real-time constraints.
We introduce a novel Bump-Image-Enhanced Voxel Representation (BIEVR) designed to explicitly encode high-resolution geometric information for efficient registration. BIEVR represents local surface geometry within each voxel as a 2D height image. Specifically, we estimate a dominant plane per voxel, which serves as an image plane, and store per-pixel deviations of observed points from this plane ("bump-image"). This representation preserves details while enabling constant-time map lookups and efficient, direct scan-to-map registration.

Although BIEVR enables high-resolution mapping, a map area can only contribute constraints to the registration if a nearby point is included in the sampled input point cloud. However, these subtle features can be missed due to downsampling of the input scan, which is often required to reduce computational cost. 
Selecting relevant points is difficult because LiDARs have limited spatial resolution, particularly at longer ranges, and geometric details can be too subtle to be detectable from the incoming scan alone. As a result, these details often emerge only after integrating multiple scans into the map.
We therefore propose a map-aware sampling strategy that leverages the proposed representation. For each voxel, we compute a score derived from its height image, that reflects its potential to provide informative registration Jacobians. Given the initial pose estimate from IMU integration, we use a higher point-density in voxels with a high score. This approach focuses computation on informative areas, while using significantly fewer points than uniform high-density sampling.

We evaluate our approach extensively across multiple LiDAR sensors and robotic platforms using a single, unified set of parameters, highlighting the broad applicability of the method. The results demonstrate that our method achieves exceptional robustness in severely uninformative environments where all compared baselines fail. It also delivers state-of-the-art performance in well-constrained scenarios at competitive runtime. Finally, beyond odometry, we demonstrate how BIEVR-LIO naturally supports downstream robotics tasks, by showing real-world experiments where the map is used for footstep planning on a legged robot.

In summary, we propose two contributions to exploit the geometric details in LiDAR data: \textit{(1)} a high-resolution voxel map in which each voxel stores an oriented bump image, and \textit{(2)} a map-informed, dual-resolution point sampling scheme. 

We open-source our implementation at \url{https://github.com/ethz-asl/bievr-lio}.

\section{Related Work}
\label{sec:related_work}

\subsection{LiDAR-based Odometry}
One of the first widely used LiDAR odometry systems was LOAM~\cite{zhang2014loam}, which uses planar and edge features for registration. KISS-ICP~\cite{vizzo2023kiss} showed that strong performance is also achievable with a simple point-to-point ICP design. While those approaches register full scans, others treat the measurements as a point stream for continuous trajectory estimation~\cite{cao2025resple, he2023point, zheng2024traj, burnett2024continuous}. IMUs enable motion compensation and provide a registration prior and are thus often fused with LiDAR. Typically, LIO systems follow a tightly-coupled fusion, using factor graphs ~\cite{koide2024glim, zhao2021super, shan2020lio} or filters~\cite{xu2022fast, he2023point, cao2025resple, bai2022faster}. Different from the preintegration~\cite{forster2016manifold} used in these systems, RKO-LIO~\cite{malladi2025robust} presents a simplified IMU model that is used to regularize registration. Contrary to existing work, the IMU does not influence the registration beyond the scan-undistortion and initialization in our loosely-coupled pipeline.

A common challenge for LIO systems is geometric degeneracy, encountered in environments with little informative structure. Several degeneracy detection methods have been proposed~\cite{tuna2023x, zhao2025superloc, lee2025lodestar}, however, they only handle short uninformative sections or require odometry from additional sensor sources. Other work addresses degeneracy by using LiDAR intensity as an additional source of constraints~\cite{pfreundschuh2024coin, khedekar2025pg, zhang2023ri}. Despite impressive results, these rely on dense LiDARs and do not support irregular or sparse scans that are often found in cheaper sensors. We instead use a high-resolution voxel-image map that increases gradient observability, combined with a map-informed point sampling strategy to prevent degeneracy.

\subsection{Map Representations}
The most direct map representation in LiDAR odometry is a point cloud map, commonly maintained using keyframes~\cite{chen2022direct}, submaps~\cite{zhao2021super}, or a globally accumulated map~\cite{xu2022fast}, which can also be continuously optimized~\cite{koide2024glim}. These maps are typically indexed using kd-tree structures, leading to $\mathcal{O}(\log N)$ point-lookup complexity. This forces methods to downsample aggressively to stay real-time, losing detail. Faster-LIO~\cite{bai2022faster} improved efficiency with $\mathcal{O}(1)$ lookups in a hashed voxel map. However, unless point-to-point registration is used~\cite{vizzo2023kiss, malladi2025robust, lee2024genz}, map points are not directly used in the registration objective. Instead, intermediate representations such as surface normals or covariances are estimated from local neighborhoods, which presents a computational bottleneck.

To avoid repeated neighborhood queries, voxel-based methods directly track a normal~\cite{yuan2022efficient} or covariance~\cite{chen2024ig} per voxel. Closely related are surfel maps~\cite{behley2018efficient, quenzel2025lio, nguyen2023slict}, which represent the environment as oriented surface elements. Although efficient, such representations assume that geometry can be approximated by a plane or a Gaussian, which limits the level of detail.

Other work directly uses distance fields for registration. D-LIO~\cite{11248856} proposes a discrete truncated distance field updated efficiently via a binary kernel, while 2Fast-2Lamaa~\cite{gentil20242fast} models a continuous distance field using Gaussian Processes, enabling smooth point-to-surface distance queries.

Motivated by map compression, CURL-SLAM~\cite{zhang2025curl} represents the surface inside a voxel using spherical harmonics, reducing the memory footprint. Points are first projected onto an axis-aligned plane to generate a height image, which is then approximated by optimizing harmonic coefficients. Because harmonic fitting is computationally expensive, map updates are not performed at LiDAR frequency, and the resulting approximation is inherently smooth, suppressing subtle geometric cues that are valuable for registration. Registration is performed against resampled height images by minimizing the difference between the projected point height and the pixel value. 
Our approach builds on the same registration paradigm, but explicitly stores voxel-wise oriented height images at high resolution, which enables efficient map updates and constant-time pixel lookups without reconstruction or neighborhood queries. We do not assume an underlying geometric primitive and preserve sharp discontinuities, allowing subtle geometric gradients to be used for robust odometry.
\begin{figure*}[]
\begin{center}
\includegraphics[width=\linewidth]{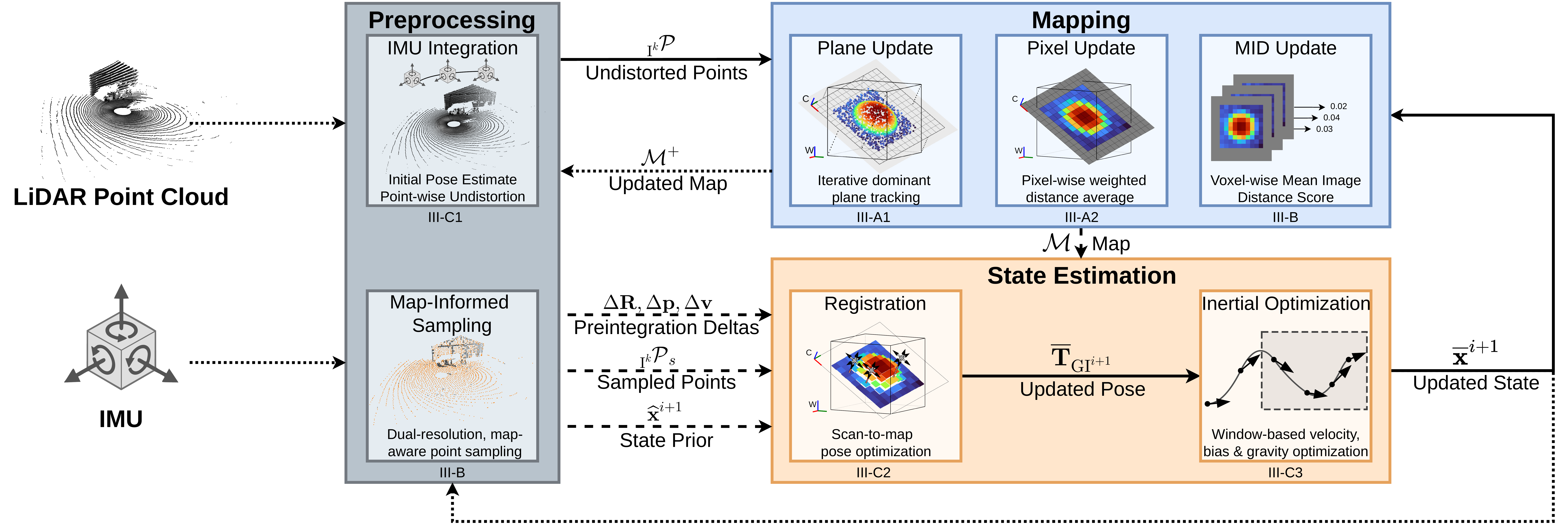}
\vspace{-4mm}
\caption{Overview of the proposed BIEVR-LIO system. Lines indicate information flow before ($\cdots$), during (- - -) and after (---) the state update.}
\label{fig:pipeline}
\vspace{-7mm}
\end{center}
\end{figure*}
\subsection{Point Sampling}
Modern 3D LiDARs produce more points than can be registered in real-time, necessitating downsampling. A widely used strategy is fixed-size voxel grid filtering, where the number of sampled points depends on the scale of the environment, which can result in unstable registration in narrow environments. To address this, AdaLIO~\cite{lim2023adalio} adapts the voxel size to the environment extent.
LOAM~\cite{zhang2014loam} instead samples specific edge and planar features.
Since informative points are usually not evenly distributed in a scan, RMS~\cite{petracek2024rms} presents a sampling scheme that minimizes redundancy within the point cloud. 
These methods sample exclusively based on the current input scan. However, degeneracy is caused by a lack of informative Jacobians in the registration objective, which also depends on the map. Consequently, other approaches~\cite{tuna2023x,jiao2021greedy, 9561324} sample based on the Hessian of the optimization problem. This requires building computationally expensive correspondences and residual terms, which then necessitates coarse downsampling before score computation, and in turn can remove crucial details. Our approach avoids the calculation of residual terms by defining a metric derived from the voxel-wise height images to approximate the potential to provide informative Jacobians, and selects more points in voxels with high potential.

\section{Method}
We propose BIEVR-LIO, a system that exploits subtle geometric details while maintaining computational efficiency. At its core, it represents surfaces as voxel-wise oriented height images, enabling storage of fine-grained geometry with constant-time lookups and efficient updates. IMU integration is used for pointwise undistortion of the LiDAR scan, and provides an initial pose estimate for registration. Our system follows a loosely-coupled design in which the pose is estimated exclusively from scan-to-map registration, without any inertial residuals. We present a map-aware sampling strategy that uses a lightweight metric to increase point density in informative areas. Registration is then performed directly on the BIEVR representation by minimizing the height error between sampled input points and their corresponding voxel height images. After pose estimation, we optimize velocity, gravity, and IMU biases. Finally, we update the map using all scan points, made tractable by an efficient, parallelized process. We provide an overview of the system in \Cref{fig:pipeline}.

\subsection{Bump-Image-Enhanced Voxel Representation}
\label{sec:map}
LIO systems often approximate surfaces as planes, which enables efficient registration constraints. In practice, however, choosing an appropriate spatial resolution for planar approximation is difficult, as large support regions suppress geometric detail, while small regions lead to unstable plane estimates.
Instead of assuming that a plane fully explains the surface, we explicitly capture the local deviation from a dominant plane per voxel. This plane acts as the image plane of a height image, in which pixels store the fine-grained deviation orthogonal to the plane. By tracking the deviation from a well-defined plane, this representation preserves subtle geometric cues that are lost under coarse planar models, while avoiding the instability of estimating high-resolution 3D primitives directly from points.

\begin{figure*}[t]
\centering
\vspace{-12mm}

\begin{minipage}[t]{0.33\textwidth}
    \centering
    \includegraphics[width=\linewidth]{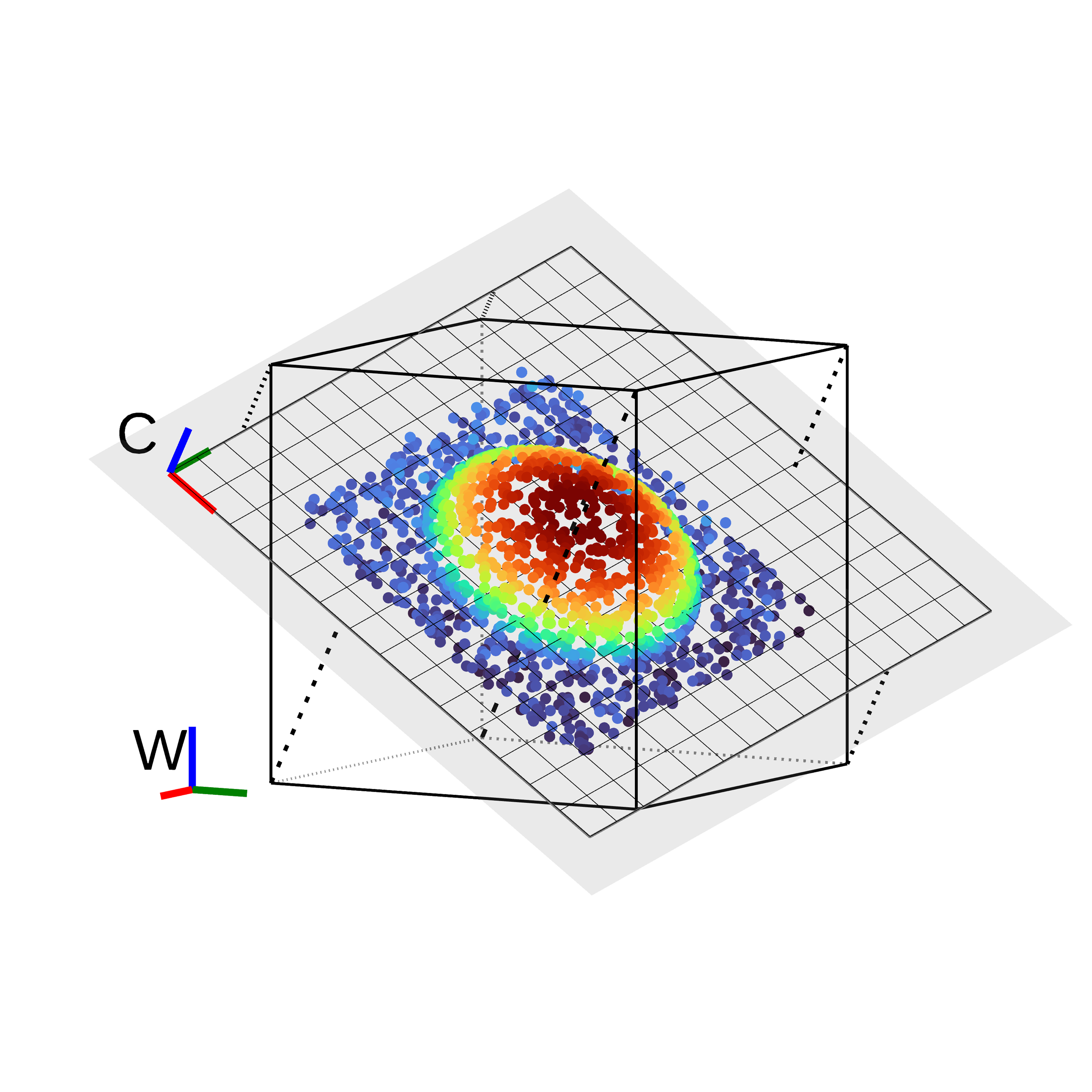}
    \vspace{-17mm}
    \par\smallskip
    \textbf{(a)} Plane update
\end{minipage}%
\begin{minipage}[t]{0.33\textwidth}
    \centering
    \includegraphics[width=\linewidth]{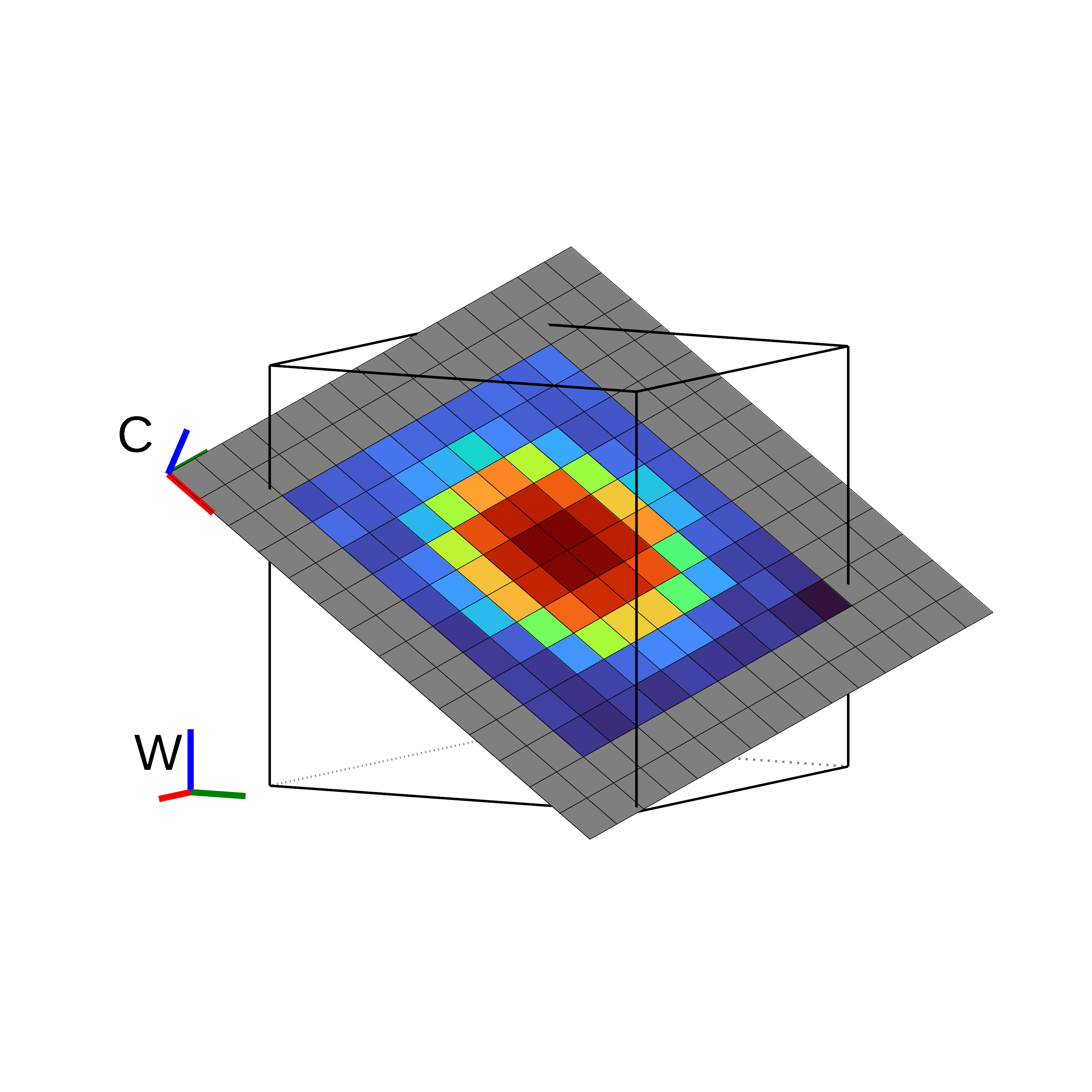}
    \vspace{-17mm}
    \par\smallskip
    \textbf{(b)} Projection
\end{minipage}%
\begin{minipage}[t]{0.33\textwidth}
    \centering
    \includegraphics[width=\linewidth]{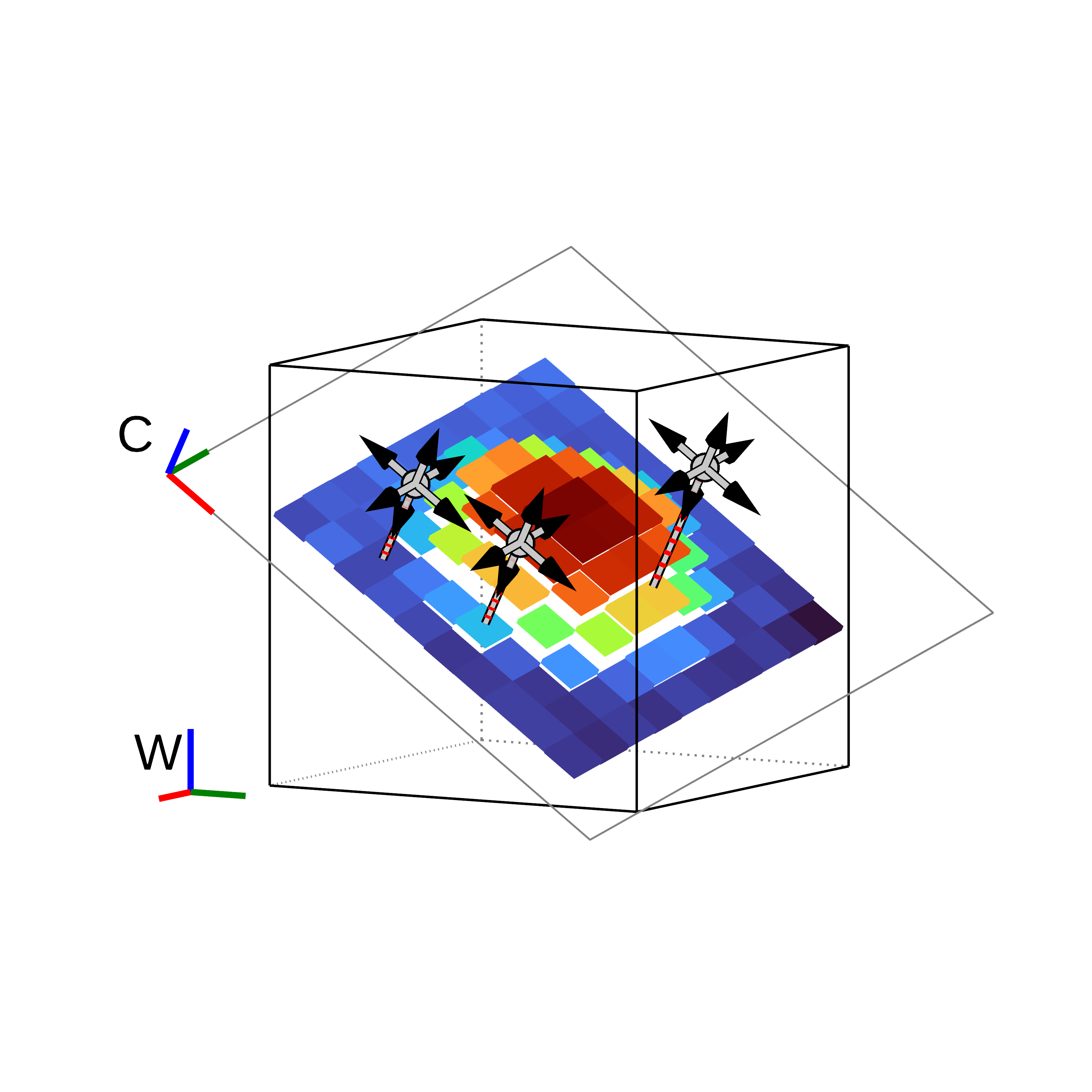}
    \vspace{-17mm}
    \par\smallskip
    \textbf{(c)} Registration
\end{minipage}

\vspace{-9.5mm}
\caption{Map update and registration. \textit{(a)} A dominant plane is iteratively updated. The image size is determined by projecting voxel corners onto this plane. Points are coloured based on plane distance. \textit{(b)} Points are projected to pixels, and pixel values are updated based on height above the plane. \textit{(c)} During registration, we minimize height residuals between pixel values and input points (gray). Arrows indicate the Jacobian components described in \Cref{eq:jacobian}.}
\label{fig:update}
\vspace{-4mm}
\end{figure*}

\subsubsection{Iterative Plane Estimation}
For the iterative plane estimation and normal update, we adopt the voxel-wise update strategy presented in iG-LIO~\cite{chen2024ig}. Undistorted LiDAR points are transformed to the world frame using the pose estimate and grouped to voxels of \SI{0.5}{\meter} length via a spatial hash function. The individual voxels are then updated in parallel.
For each voxel $\mathcal{V}$, we update the sum of points $\mathbf{s}$, the sum of outer products $\mathbf{C}$, and the number of points $n$ as
\vspace{-0.5mm}
\begin{equation}
\mathbf{s} \leftarrow \mathbf{s} + \sum_{p_i\in\nu}^{} \mathbf{p}_i, \quad
\mathbf{C} \leftarrow \mathbf{C} + \sum_{p_i\in\nu}^{} \mathbf{p}_i \mathbf{p}_{i}^\top, \quad
n \leftarrow n + n_\nu,
\end{equation}
where $\nu$ is the set of points $\mathbf{p}_i$ falling into $\mathcal{V}$, and $n_\nu = |\nu|$ is the number of points added to that voxel from the current scan.
The centroid and covariance can then be extracted as:
\vspace{-0.5mm}
\begin{equation}
\boldsymbol{\mu} = \frac{\mathbf{s}}{n}, \qquad \boldsymbol{\Sigma} = \frac{\mathbf{C} - \mathbf{s} \boldsymbol{\mu}^\top }{n + 1} .
\end{equation}
Based on eigenvalue decomposition of $\boldsymbol{\Sigma}$, we extract the voxel normal as the eigenvector with the smallest eigenvalue. We create a coordinate system with origin at $\boldsymbol{\mu}$ and set the z-axis as the normal. We define the transform\footnote{Throughout this paper, we use the shorthand ${}_\mathrm{A}\mathbf{p} = \mathbf{T}_\mathrm{AB}\,{}_\mathrm{B}\mathbf{p}$ to denote the homogeneous transformation $[{}_\mathrm{A}\mathbf{p}^\top, 1]^\top = \mathbf{T}_\mathrm{AB} \, [{}_\mathrm{B}\mathbf{p}^\top, 1]^\top$} $\mathbf{T}_{\mathrm{VG}}\in \rm{SE(3)}$, with $\mathbf{R}_{\rm{VG}} \in \rm{SO(3)}$ its rotational component, that maps points from the global world frame $\mathrm{G}$ to the voxel coordinate frame $\mathrm{V}$.

\subsubsection{Bump Image Update}
Each voxel in the BIEVR map stores an image $\mathbf{I}$ representing fine-grained surface deviations perpendicular to the local plane. 
We initialize a voxel once at least four points were accumulated in it. Upon initialization we determine the minimal required image size to represent the full volume of the voxel by projecting its eight corners onto the $x$-$y$ plane of the voxel coordinate system, as shown in \Cref{fig:update}. The image width and height are computed as 
\begin{equation}
W = \frac{x_{\max} - x_{\min}}{r}, \quad H = \frac{y_{\max} - y_{\min}}{r},
\end{equation}
where $r$ is the chosen pixel resolution and $x_{\min}$, $x_{\max}$, $y_{\min}$, $y_{\max}$ are the extrema of the projected corners. 
We define a height image coordinate system $\mathrm{C}$ with $\mathbf{R}_{\mathrm{GC}} = \mathbf{R}_{\mathrm{GV}}$ and origin at ${}_\mathrm{G}\mathbf{p}_\mathrm{C} = \mathbf{T}_{\mathrm{GV}} [x_{\min}, y_{\min}, 0]^\top$.

To calculate the height image values (\Cref{fig:update}), we project each input point in a voxel to its image coordinates: 

\begin{equation}
{}_\mathrm{C}\mathbf{p}_i = \mathbf{T}_{\mathrm{CG}} \,{}_G\mathbf{p}_i, \quad
u = \frac{[{}_\mathrm{C}\mathbf{p}_i]_x}{r}, \quad v = \frac{[{}_\mathrm{C}\mathbf{p}_i]_y}{r}, \quad z = [{}_\mathrm{C}\mathbf{p}_i]_z
\end{equation} where $[{}_\mathrm{C}\mathbf{p}_i]_x$, $[{}_\mathrm{C}\mathbf{p}_i]_y$, $[{}_\mathrm{C}\mathbf{p}_i]_z$ denote the individual coordinates of ${}_\mathrm{C}\mathbf{p}_i$.
To accumulate information from multiple scans in the bump image $\mathbf{I}$, we update pixel values with a weighted averaging. As we use a small pixel size of \SI{0.05}{\meter}, it is inevitable that points occasionally fall into the nearby pixels. As orientation errors result in larger Euclidean error for points further away from the LiDAR, we regard points close to the LiDAR as more reliable and assign them a higher weight: 

\begin{flalign}
&\mathbf{I}(u,v) \leftarrow \frac{\mathbf{W}(u,v) \cdot \mathbf{I}(u,v) + w_i \cdot z_i}{\mathbf{W}(u,v) + w_i}  \\
&\mathbf{W}(u,v) \leftarrow \mathbf{W}(u,v) + w_i,
\quad w_i = \min(0.5,\|{}_\mathrm{L}\mathbf{p}_i\|^{-1})
\end{flalign}
where $\mathbf{W}$ stores the pixel weights.
After updating, we apply a Gaussian filter with 1 pixel radius to mitigate pixel noise.

If the voxel normal changes by more than $3^\circ$, we update the image coordinate system and recompute the image by reprojecting the previous image from the old coordinate frame $\mathrm{C}_o$ to the new plane frame $\mathrm{C}_n$:

\begin{equation}
{}_{\mathrm{C}_n}\mathbf{p} = \mathbf{T}_{\mathrm{C}_n\mathrm{C}_o} 
\big[u \cdot r,  v \cdot r, \mathbf{I}_o(u,v) \big]^\top, 
\end{equation}

\begin{equation}
\mathbf{I}_n\Big(\frac{[{}_{\mathrm{C}_n}\mathbf{p}_i]_x}{r},\frac{[{}_{\mathrm{C}_n}\mathbf{p}_i]_y}{r}\Big) = [{}_{\mathrm{C}_n}\mathbf{p}_i]_z
\end{equation}

As this map update is computationally cheap, we are able to integrate the full undistorted input scan, which provides more information than downsampled points used in other works.

\subsubsection{Implementation}
We implement BIEVR-LIO in a hash-map, where voxels are indexed by their spatial location using Morton code~\cite{morton1966computer}, which achieves $\mathcal{O}(1)$ average-case lookup. Each voxel stores the transform $\mathbf{T}_\mathrm{CG}$, the height image $\mathbf{I}$, pixel weights $\mathbf{W}$, the sum of points $\mathbf{s}$, the sum of outer products $\mathbf{C}$, the number of included points $n$ and its Mean Image Distance (\Cref{sec:sampling}). Voxels are efficiently updated in parallel.

To enforce bounded memory, we maintain a Least Recently Used cache. If the number of stored voxels exceeds a configured capacity, the least recently updated voxels are evicted.

\subsection{Map-Informed Point Sampling}
\label{sec:sampling}
The fine-grained geometric detail captured in BIEVR can only contribute to registration if a corresponding input point is sampled in its vicinity. Uniformly sampling the input scan at high density, however, increases the registration's computational cost and introduces noise from uninformative regions. To address this, we propose a map-aware, two-stage sampling strategy that prioritizes points in informative areas.

We first quantify the potential of a voxel to provide useful gradients beyond its planar component. Conveniently, the height image itself is an indicator of non-planarity. Flat surfaces result in small deviations from the voxel plane, translating into small pixel distance, while non-planar surfaces cause large values. Thus, we use the \ac{MID}, calculated over observed pixels, as an indicator value:
\vspace{-2.5mm}
\begin{equation}
\text{MID} = \frac{1}{|\Omega_\text{obs}|} \sum_{(u,v) \in \Omega_\text{obs}} \|\mathbf{I}(u,v)\|,
\vspace{-1mm}
\end{equation}
with $ \Omega_\text{obs} = \{(u,v) : \mathbf{W}(u,v) > 0\}$.
Voxels with larger \ac{MID} contain richer geometric detail and are more likely to contribute informative Jacobians. This computationally efficient metric is updated during mapping, which avoids additional computation during the sampling procedure.

\begin{figure}[b]
\vspace{-4mm}
\includegraphics[width=\linewidth]{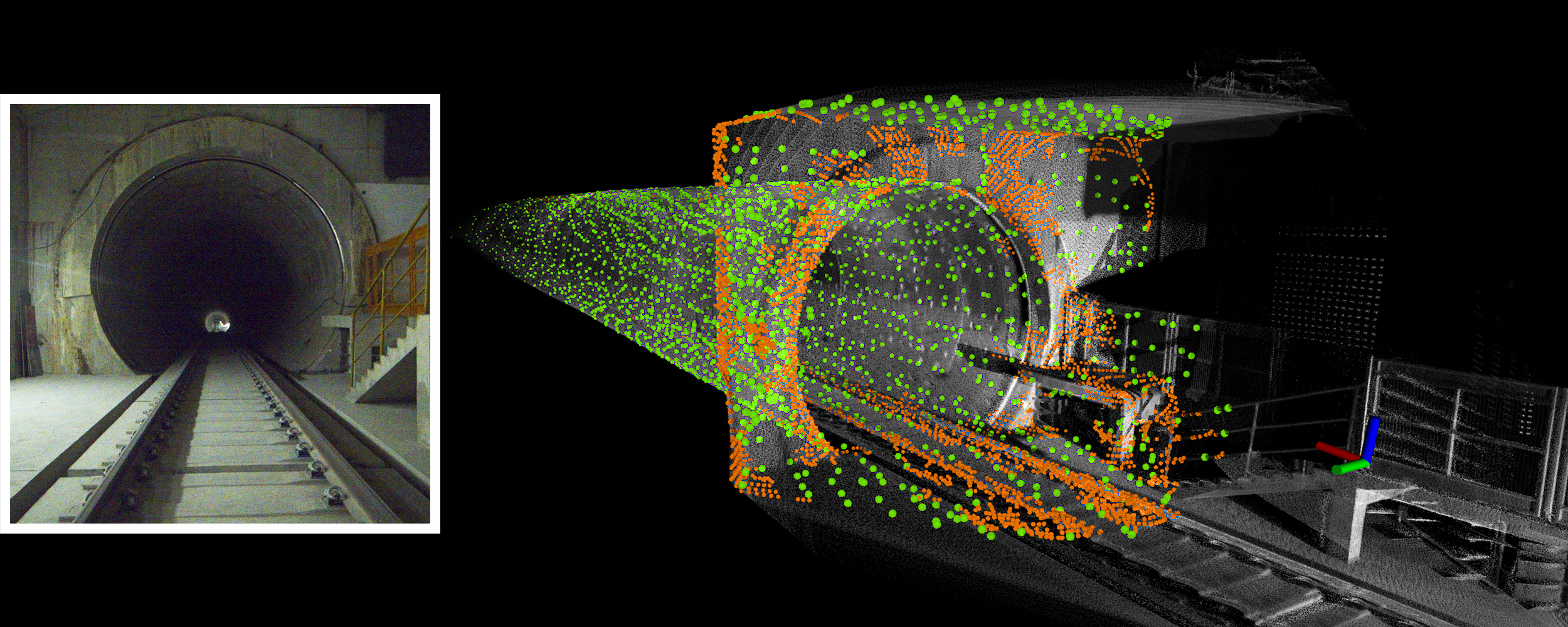}
\vspace{-7mm}
\caption{Point sampling in \textit{Shield1}. Gray points show an accumulated point map. Based on the \ac{MID} metric, dense points (orange) are sampled in salient regions such as tunnel entrance corners and the track bed, while sparse points (green) are sampled in less informative regions such as the ceiling and floor.}
\label{fig:sampling}
\end{figure}

The sampling proceeds in two stages. First, the undistorted point cloud is downsampled at a high resolution (\SI{0.1}{\meter}) and transformed into the world frame using the IMU-based initial pose estimate. For each downsampled point, we query its corresponding map voxel and select the 300 voxels with the highest \ac{MID} to retain all points within them. Second, the points in the remaining voxels are further downsampled at a coarser resolution (\SI{0.5}{\meter}). The combination of high-resolution points from informative voxels and low-resolution points elsewhere (shown in \Cref{fig:sampling}) is then used for registration.

\label{sec:method:sampling}

\subsection{Loosely-Coupled State Estimation}
\label{sec:loosely_coupled}

BIEVR-LIO estimates the robot state defined as\\ 
$\mathbf{x} \triangleq [
\mathbf{R}_\mathrm{GI} \;\,
{}_\mathrm{G}\mathbf{t}_\mathrm{GI} \;\,
{}_\mathrm{G}\mathbf{v}_\mathrm{I} \;\,
\mathbf{b}_a \;\,
\mathbf{b}_g \;\,
{}_\mathrm{G}\mathbf{g}] 
\in SO(3) \times \mathbb{R}^{12} \times S^2
$ \\
where $\mathbf{R}_\mathrm{GI}$ denotes the rotation from the IMU frame $\mathrm{I}$ to the global frame $\mathrm{G}$, ${}_\mathrm{G}\mathbf{t}_\mathrm{GI}$ and ${}_\mathrm{G}\mathbf{v}_\mathrm{I}$ are the IMU position and velocity expressed in the global frame, $\mathbf{b}_a$ and $\mathbf{b}_g$ represent the accelerometer and gyroscope biases, and ${}_\mathrm{G}\mathbf{g}$ is the gravity direction in the global frame.

The majority of LIO systems follow a tightly-coupled formulation, motivated by findings~\cite{ye2019tightly} that IMU measurements can temporarily compensate for missing geometric constraints in short degenerate segments. 

However, tight coupling introduces a weighting between modalities, requiring tuning of sensor specific noise parameters. In extended low-structure regions, we observed that inertial residuals can dominate the optimization, potentially overwhelming subtle but informative registration cues. As our work specifically aims to exploit such geometric detail, we propose a loosely-coupled strategy that preserves the IMU's benefits in motion compensation and registration initialization, while isolating the pose estimation from inertial residual influence.
Specifically, $\mathbf{T}_\mathrm{GI}$ is computed only from registration (\Cref{sec:registration}), without any inertial residuals or covariances, and the remainder of the state is optimized separately (\ref{sec:window}).

\subsubsection{IMU-Based State Prediction \& Motion Compensation}
\label{sec:imu_prediction}

We adopt the piecewise linear continuous-time model of~\cite{le2023continuous} to compute preintegrated rotation and translation increments~\cite{forster2016manifold} between the IMU timestamps $t_j$ and $t_k$ associated with two consecutive LiDAR scans,
yielding $\Delta \mathbf{R}_{j,k} \in \mathrm{SO}(3)$ and $\Delta \mathbf{p}_{j,k} \in \mathbb{R}^3$. These increments are combined with the previous state to form $\mathbf{T}_{\mathrm{I}^k \mathrm{I}^j}$, which is used to undistort a LiDAR points from $t_i$ to the most recent timestamp $t_k$.

During this step, points are also transformed to the IMU frame for downstream processing: ${}_{\mathrm{I}^k}\mathbf{p}_i = \mathbf{T}_{\mathrm{I}^k\mathrm{I}^i} \mathbf{T}_{\mathrm{I}^i\mathrm{L}^i} \, {}_{\mathrm{L}^i}\mathbf{p}_i$,
where $\mathbf{T}_{\mathrm{I}^i\mathrm{L}^i}$ denotes the calibration from the LiDAR frame to the IMU frame.
The same increments are used to predict a registration prior $\mathbf{T}_{\mathrm{GI}^k}$. Finally, they also define the inertial residuals in the optimization described in Sec.~\ref{sec:loosely_coupled}.

\subsubsection{Registration}
\label{sec:registration}
We align the sampled point cloud directly to the map by minimizing the error between the projected heights of input points and the value of their respective pixel in the voxel height-map. This process is inspired by CURL-SLAM~\cite{zhang2025curl} and can be thought of as a photometric minimization in a virtual image, where the pixel values represent the height above the voxel image plane, as shown in \Cref{fig:update}.
Each input point ${}_\mathrm{I}\mathbf{p}_i$ is mapped to a voxel by hashing its coordinates based on the initial pose estimate. The residual is defined as:
\vspace{-3mm}
\begin{equation}
\begin{split}
r_i(\boldsymbol{\xi}) =[{}_\mathrm{C}\mathbf{p}_{i}]_z(\boldsymbol{\xi}) - \mathbf{I}\big(u_i(\boldsymbol{\xi}), v_i(\boldsymbol{\xi})\big) \\
{}_\mathrm{C}\mathbf{p}_i(\boldsymbol{\xi}) = \mathbf{T}_{\mathrm{CG}} \mathbf{T}_{\mathrm{GI}} \mathrm{Exp}(\boldsymbol{\xi}) \, {}_\mathrm{I}\mathbf{p}_i
\end{split}
\vspace{-3mm}
\end{equation}
where $[{}_\mathrm{C}\mathbf{p}_{i}]_z$ is the point height above the image plane, $(u_i, v_i)$ are the pixel coordinates in that voxel's height image and $\mathrm{Exp}(\boldsymbol{\xi})$ maps the pose increment $\boldsymbol{\xi} \in \mathfrak{se}(3)$ to a transformation $\mathbf{T} \in SE(3)$. The image value $\mathbf{I}(u_i, v_i)$ is determined by bilinear pixel interpolation among observed pixels. If a point projects to a region where all four surrounding pixels are unobserved, it is omitted.

The Jacobian of the residual with respect to the pose perturbation $\boldsymbol{\xi}$ can be divided into two components:
\begin{equation}
\frac{\partial r_i}{\partial \boldsymbol{\xi}} =  
\underbrace{\frac{\partial [{}_\mathrm{C}\mathbf{p}_{i}]_z}{\partial \boldsymbol{\xi}}}_{\textit{I}} - \underbrace{\frac{\partial \mathbf{I}(u_i, v_i)}{\partial \boldsymbol{\xi}}}_{\textit{II}} 
\label{eq:jacobian}
\end{equation}
We provide a full expansion of (\ref{eq:jacobian}) in the supplementary material and want to highlight the geometric interpretation of the two Jacobian components. (I) can be understood as a normal-direction term, equivalent to classical point-to-plane error behavior, and describes how the residual changes as the point moves towards the image plane. Orthogonal to this, the second term (II) provides two Jacobian directions derived from the height image gradient, which describe the residual change from moving the point parallel to the plane, i.e., how the image value changes if the point projects to neighboring pixels. These two additional directions help to augment the optimization with information in directions that are unobservable under point-to-plane alignment, which increases robustness in uninformative regions, as shown in our experiments (\Cref{sec:ablation}).

We solve the resulting non-linear least-squares problem using Levenberg-Marquardt minimization:
\vspace{-1mm}
\begin{equation}
\boldsymbol{\xi}^* = \underset{\boldsymbol{\xi}}{\arg\min} \sum_i \rho\Big(\big\| r_i(\boldsymbol{\xi}) \big\|\Big),
\vspace{-1mm}
\end{equation}
where \(\rho(\cdot)\) denotes the Huber loss function.

\subsubsection{Sliding-Window Inertial Optimization}
\label{sec:window} 
While poses are determined by LiDAR registration, we incorporate IMU measurements via a sliding-window optimization to enforce temporal inertial consistency using IMU residuals $\bm{\mathit{r}}_{\!IMU}$ as in~\cite{qin2018vins}. 
Since gravity direction and IMU biases evolve slowly~\cite{el2008analysis}, we estimate them as window-constant variables rather than per-timestep states. 
All poses within the \SI{10}{\second} window are fixed and serve only as anchors, yielding the following optimization:
\begin{equation}
\begin{aligned}
&[\overline{\mathbf{v}}_{k-n}, \dots, \overline{\mathbf{v}}_k],\overline{\mathbf{g}},\, \overline{\mathbf{b}}_a,\overline{\mathbf{b}}_g = \\
& \!\arg\min\!\!\!\!\sum_{i=k-n}^{k-1} \!\!\!
\Big\| \bm{\mathit{r}}_{\!IMU}\!\Big(\!
\mathbf{v}_i, \!\mathbf{v}_{i+1}, \mathbf{g}, \mathbf{b}_a, \!\mathbf{b}_g, 
\underbrace{\mathbf{T}_{\mathrm{GI^i}},\! \mathbf{T}_{\mathrm{GI^{i+1}}} }_{\text{fixed}}
\Big)\!\Big\|^2
\end{aligned}
\vspace{-1.5mm}
\end{equation}
This preserves sensitivity to geometric details while estimating biases and avoiding covariance tuning, which improves robustness across sensor configurations and environments.

\section{Experiments}
\label{sec:experiments}

\begin{table}[t]
 \caption[]{Parameter Overview}
 \vspace{-2mm}
\label{tab:parameters}
\centering
\begin{tabular}{lcc}
\toprule
\textbf{Parameter} & \textbf{Symbol} & \textbf{Value} \\
\midrule
Voxel side length & $l$ & \SI{0.5}{\meter} \\
Bump-Image resolution & $r$ & \SI{0.05}{\meter} \\
Downsample resolution coarse / fine & $d_c$ / $d_f$ & 0.5 / \SI{0.1}{\meter} \\
\# Fine-resolution voxels & --- & 300 \\
Huber Loss Delta & $\delta$ & \SI{0.1}{\meter} \\
IMU Optimization Window & --- & \SI{10}{\second} \\
\bottomrule
\end{tabular}
\vspace{-3mm}
\end{table}

We evaluate our system against a broad set of state-of-the-art LiDAR–based odometry methods. Specifically, we compare against KISS-ICP~\cite{vizzo2023kiss}, GenZ-ICP~\cite{lee2024genz}, Traj-LO~\cite{zheng2024traj}, FAST-LIO2~\cite{xu2022fast}, DLIO~\cite{chen2022direct}, iG-LIO~\cite{chen2024ig}, RESPLE~\cite{cao2025resple} and RKO-LIO~\cite{malladi2025robust}. Furthermore, for datasets providing high-resolution scans (\Cref{tab:ncd},\ref{tab:enwide}), we also include COIN-LIO~\cite{pfreundschuh2024coin}, which additionally uses LiDAR intensity information.
For CURL-SLAM~\cite{zhang2025curl}, we only list the results (\Cref{tab:ncd}) from the publication, as the code was not publicly available at the time of submission.
To assess BIEVR-LIO's generalization capabilities, we evaluate on multiple publicly available datasets spanning different platforms, LiDAR configurations, and environments. These datasets include geometrically rich scenes as well as challenging environments with limited structural information to analyze performance across a wide range of applications.
We use the same parameters, listed in \Cref{tab:parameters}, for all experiments, datasets, and sensors without any per-scene or per-dataset tuning, which underlines the robustness of our method.
For the baseline methods, we use sensor-specific parameter configurations when available and, otherwise, rely on the publicly available default parameters for each approach. 

We evaluate the global trajectory accuracy using the Absolute Trajectory Error (ATE) and the relative error (RE) over trajectory segments of \SI{10}{\meter} as an estimate of the local drift. Since the evaluated datasets only contain position ground truth, we calculate the RE using the point distance metric\footnote{\url{https://github.com/MichaelGrupp/evo}\label{fn:evo}}. Trajectory alignment and evaluation are performed using evo\footref{fn:evo}. Methods with relative error exceeding 20\% are considered to have failed ($\times$), and their ATE is not reported, as excessive drift prevents meaningful alignment to the ground truth.
We conduct all experiments on an Intel i7-11800H CPU.

\begin{table}[bh!]
\caption{Newer College Dataset Results}
\label{tab:ncd}
\vspace{-2mm}
\begin{adjustbox}{max width=\linewidth}
\begin{tabular}{lcccc}
\toprule
\multicolumn{5}{c}{Absolute Trajectory Error RMSE (\SI{}{\meter}) / Relative Error RMSE (\textit{\%})} \\
\midrule
Method & QuadHard & Cloister & Stairs & Park \\
\midrule
KISS-ICP & $0.324$ / $1.9$ & $0.297$ / $2.1$ & $\times$ / $32.5$ & $2.871$ / $1.1$ \\
GenZ-ICP & $\times$ / $28.0$ & $0.137$ / $0.7$ & $0.948$ / $7.2$ & $0.718$ / $0.8$ \\
Traj-LO & $0.133$ / $1.0$ & $0.151$ / $0.6$ & $2.043$ / $9.3$ & $0.503$ / $0.6$ \\
CURL-SLAM & $-$ / $-$ & $0.212$ / $-$ & $0.169$ / $-$ & $\underline{0.271}$ / $-$ \\
FAST-LIO2 & $\underline{0.049}$ / $\mathbf{0.3}$ & $\underline{0.078}$ / $\underline{0.2}$ & $\times$ / $3497.2$ & $0.31$ / $0.6$ \\
DLIO & $0.14$ / $1.0$ & $0.162$ / $0.8$ & $0.122$ / $1.0$ & $0.335$ / $0.7$ \\
iG-LIO & $0.079$ / $0.5$ & $\times$ / $88.5$ & $\times$ / $81.4$ & $\mathbf{0.201}$ / $\mathbf{0.5}$ \\
RESPLE & $0.06$ / $\mathbf{0.3}$ & $0.104$ / $0.3$ & $0.194$ / $2.2$ & $0.994$ / $0.7$ \\
RKO-LIO & $1.111$ / $0.6$ & $1.03$ / $2.4$ & $4.477$ / $10.8$ & $0.889$ / $0.6$ \\
COIN-LIO & $\mathbf{0.046}$ / $0.3$ & $\underline{0.078}$ / $0.3$ & $\underline{0.102}$ / $\underline{0.7}$ & $0.287$ / $\underline{0.5}$ \\
BIEVR-LIO & $0.051$ / $0.3$ & $\mathbf{0.053}$ / $\mathbf{0.2}$ & $\mathbf{0.056}$ / $\mathbf{0.6}$ & $0.798$ / $0.6$ \\
\bottomrule
\end{tabular}
\end{adjustbox}
\vspace{-3mm}
\end{table}

\subsection{Odometry Results}

\begin{table*}[t]
\caption{ENWIDE Dataset Results}
\label{tab:enwide}
\vspace{-4mm}
\begin{center}
\begin{tabular}{lcccccccc}
\toprule
\multicolumn{9}{c}{Absolute Trajectory Error RMSE (\SI{}{\meter}) / Relative Error RMSE (\textit{\%})} \\
\midrule
Method & IntersectionS & IntersectionD & RunwayS & RunwayD & FieldS & FieldD & KatzenseeS & KatzenseeD \\
\midrule
FAST-LIO2 & $12.473$ / $29.3$ & $23.8$ / $28.1$ & $\times$ / $53.6$ & $\times$ / $59.8$ & $\underline{0.163}$ / $\underline{0.6}$ & $9.209$ / $16.1$ & $1.122$ / $4.3$ & $1.02$ / $2.4$ \\
DLIO & $\mathbf{0.177}$ / $\mathbf{0.7}$ & $3.394$ / $15.6$ & $\times$ / $48.6$ & $\times$ / $41.2$ & $0.17$ / $0.6$ & $0.617$ / $3.5$ & $\mathbf{0.188}$ / $\underline{0.5}$ & $\underline{0.328}$ / $1.7$ \\
iG-LIO & $\times$ / $53.2$ & $\times$ / $60.4$ & $\times$ / $61.6$ & $\times$ / $70.3$ & $\times$ / $46.8$ & $\times$ / $49.3$ & $\times$ / $60.8$ & $2.13$ / $14.2$ \\
RESPLE & $\times$ / $72.8$ & $\times$ / $79.6$ & $\times$ / $66.0$ & $\times$ / $68.7$ & $\times$ / $27.0$ & $\times$ / $64.5$ & $\times$ / $51.5$ & $\times$ / $114.9$ \\
RKO-LIO & $\times$ / $33.0$ & $\times$ / $38.5$ & $\times$ / $43.4$ & $\times$ / $48.0$ & $\times$ / $50.2$ & $\times$ / $36.1$ & $\times$ / $19.1$ & $\times$ / $48.7$ \\
COIN-LIO & $0.466$ / $1.2$ & $\underline{1.912}$ / $\underline{1.7}$ & $\underline{1.033}$ / $\mathbf{1.9}$ & $\mathbf{2.437}$ / $\mathbf{3.0}$ & $0.232$ / $0.8$ & $\underline{0.581}$ / $\underline{1.8}$ & $0.412$ / $1.0$ & $0.592$ / $\underline{1.6}$ \\
BIEVR-LIO & $\underline{0.231}$ / $\mathbf{0.7}$ & $\mathbf{0.404}$ / $\mathbf{0.8}$ & $\mathbf{0.44}$ / $\underline{2.2}$ & $\underline{4.35}$ / $\underline{10.8}$ & $\mathbf{0.159}$ / $\mathbf{0.3}$ & $\mathbf{0.174}$ / $\mathbf{0.8}$ & $\underline{0.194}$ / $\mathbf{0.3}$ & $\mathbf{0.243}$ / $\mathbf{0.6}$ \\
\bottomrule
\end{tabular}
\end{center}
\vspace{-2mm}
\end{table*}
\subsubsection{Newer College Dataset}
We use the Newer College Dataset~\cite{zhang2021multi} to evaluate performance in geometrically informative environments and present the results in \Cref{tab:ncd}. All sequences are captured using a handheld Ouster OS0-128 LiDAR with an integrated IMU. The \textit{Cloister}, \textit{Park}, and \textit{Quad-Hard} sequences contain rich structural detail, while the \textit{Stair} sequence is a narrow staircase with constrained geometry. 
Overall, LiDAR-only approaches show lower performance due to aggressive rotations and abrupt motions in the sequences. In contrast, BIEVR-LIO handles sudden motions robustly, as shown on the Quad-Hard sequence, which has particularly abrupt rotations. This demonstrates that our loosely-coupled design successfully exploits the advantages of IMU undistortion and registration priors.
BIEVR-LIO achieves the lowest errors on the \textit{Cloister} and \textit{Stairs} sequences. While most methods struggle on the narrow \textit{Stairs}, our approach effectively uses geometric details even in confined spaces. 
\begin{table*}[]
\caption{GEODE Dataset Results}
\label{tab:geode}
\vspace{-4mm}
\begin{center}
\begin{tabular}{lccccccc}
\toprule
\multicolumn{8}{c}{Absolute Trajectory Error RMSE (\SI{}{\meter}) / Relative Error RMSE (\textit{\%})} \\
\midrule
Method & Offroad 4 & Offroad 5 & Shield 1 & Shield 4 & Shield 5 & Tunneling 1 & Tunneling 2 \\
\midrule
GenZ-ICP & $\times$ / $38.6$ & $\times$ / $23.9$ & $\times$ / $\underline{44.2}$ & $\times$ / $\underline{37.5}$ & $\times$ / $\underline{47.9}$ & $3.61$ / $13.5$ & $2.775$ / $13.0$ \\
Traj-LO & $0.241$ / $5.7$ & $\times$ / $33.9$ & $\times$ / $524.4$ & $\times$ / $406.8$ & $\times$ / $173.2$ & $\mathbf{0.339}$ / $\mathbf{1.9}$ & $2.793$ / $19.3$ \\
FAST-LIO2 & $\underline{0.138}$ / $\underline{1.0}$ & $\underline{0.142}$ / $\underline{0.8}$ & $\times$ / $93.0$ & $\times$ / $78.8$ & $\times$ / $88.3$ & $\mathbf{0.339}$ / $2.0$ & $0.576$ / $4.3$ \\
DLIO & $2.188$ / $13.1$ & $13.507$ / $14.2$ & $\times$ / $68.5$ & $\times$ / $95.2$ & $\times$ / $63.9$ & $2.542$ / $9.9$ & $\times$ / $59.0$ \\
iG-LIO & $1.648$ / $4.8$ & $0.156$ / $0.8$ & $\times$ / $66.3$ & $\times$ / $66.1$ & $\times$ / $396.4$ & $0.345$ / $\underline{1.9}$ & $\times$ / $35.5$ \\
RESPLE & $\times$ / $53.0$ & $\times$ / $51.8$ & $\times$ / $393.1$ & $\times$ / $176.0$ & $\times$ / $147.2$ & $0.45$ / $4.8$ & $0.28$ / $\underline{2.6}$ \\
RKO-LIO & $2.142$ / $5.1$ & $0.204$ / $0.9$ & $\times$ / $64.8$ & $\times$ / $54.1$ & $\times$ / $63.7$ & $0.342$ / $2.1$ & $\underline{0.23}$ / $2.8$ \\
BIEVR-LIO & $\mathbf{0.085}$ / $\mathbf{0.7}$ & $\mathbf{0.116}$ / $\mathbf{0.7}$ & $\mathbf{0.256}$ / $\mathbf{2.1}$ & $\mathbf{0.275}$ / $\mathbf{2.4}$ & $\mathbf{0.146}$ / $\mathbf{1.3}$ & $0.34$ / $\underline{1.9}$ & $\mathbf{0.096}$ / $\mathbf{1.2}$ \\
\bottomrule
\end{tabular}
\end{center}
\vspace{-7.5mm}
\end{table*}

\subsubsection{ENWIDE Dataset}
The ENWIDE~\cite{pfreundschuh2024coin} dataset is used to evaluate performance in environments with weak geometric structure. It consists of handheld outdoor sequences using an Ouster OS0-128 with an integrated IMU with both smooth and highly dynamic motions.

\begin{figure}[!b]
\vspace{-1mm}
\includegraphics[width=\linewidth]{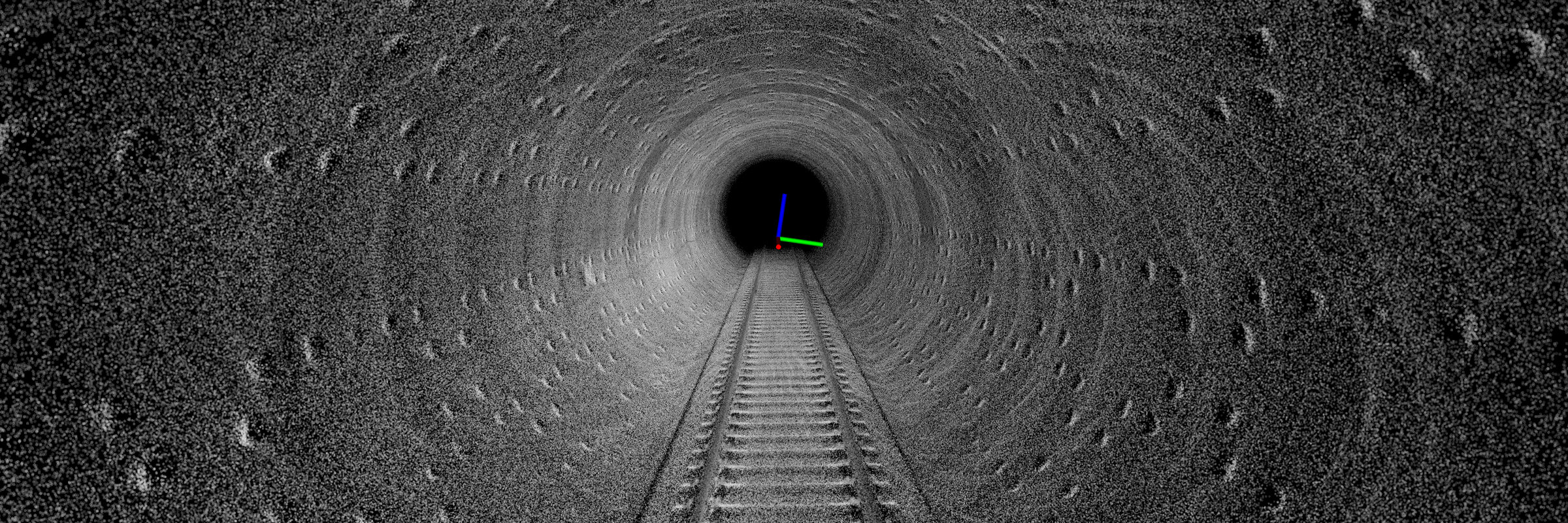}

\vspace{1mm}

\includegraphics[width=\linewidth]{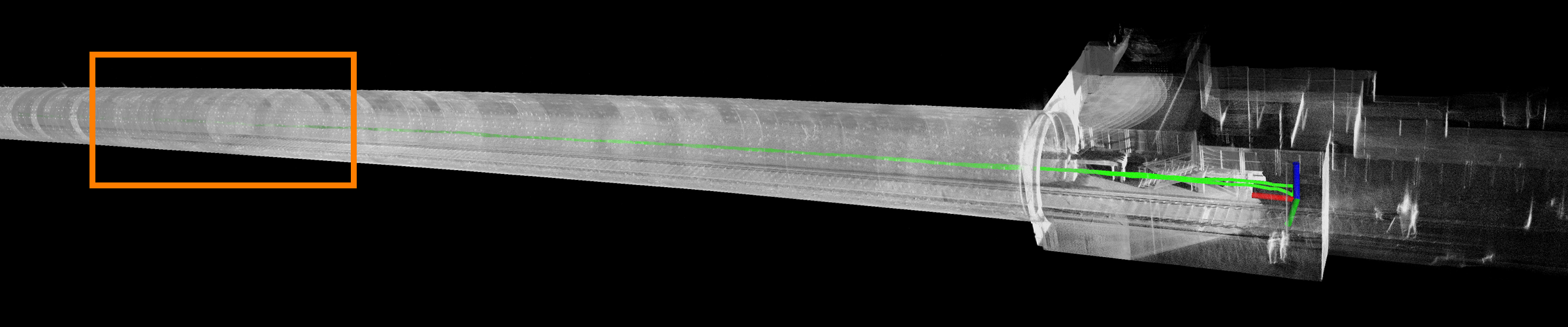}

\vspace{1mm}

\includegraphics[width=\linewidth]{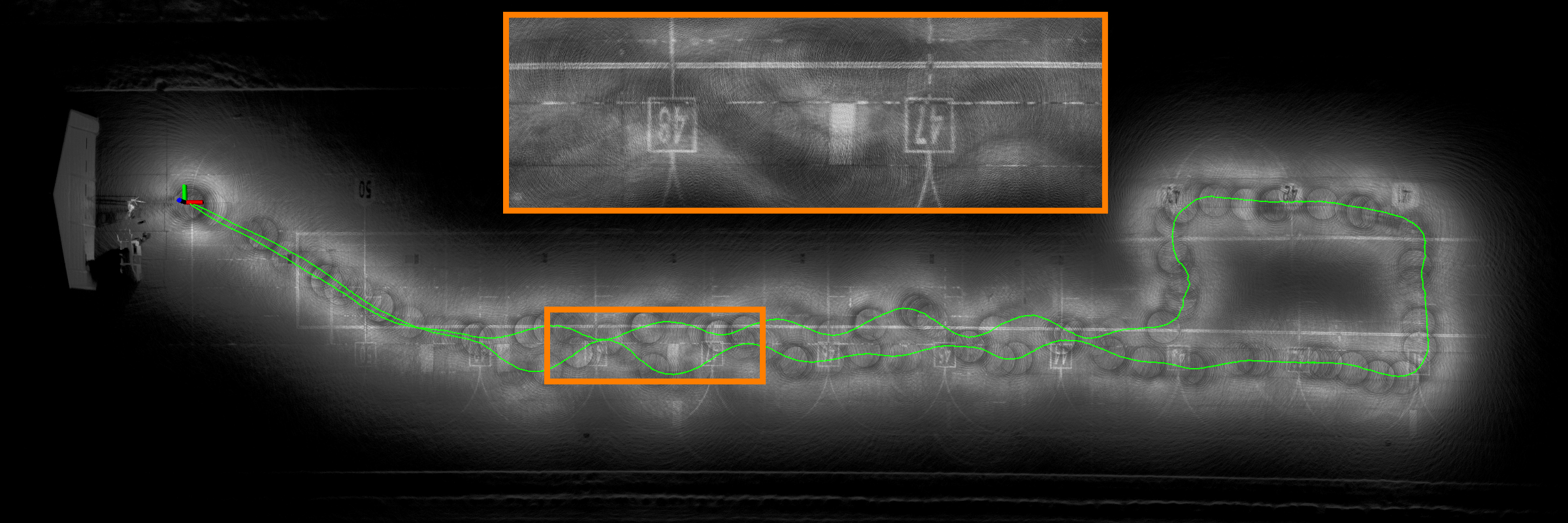}
\vspace{-7mm}
\caption{Resulting point cloud maps. \textit{Top:} Inside view of the \textit{Shield1} sequence in \textit{Middle}. The accumulated points clearly visualize the railbed and the cutouts in the tunnel wall. \textit{Middle:} Despite the challenging geometry, the map displays no visible drift after the \SI{700}{\meter} trajectory. \textit{Bottom:} After a loop around the \textit{RunwayS}, the ground markings remain crisp, illustrating the pose accuracy.}
\label{fig:mapping}
\end{figure}
Due to the challenging geometry, KISS-ICP, GenZ-ICP, Traj-LO, RESPLE, and RKO-LIO diverged on all sequences and are not included in the results in \Cref{tab:enwide}. Similarly, iG-LIO produced results only for the \textit{KatzenseeD} sequence, while FAST-LIO2 performed well on \textit{FieldS} but drifts elsewhere. DLIO achieves accurate results on the slower sequences but drifts or diverges in dynamic sequences and the extremely sparse \textit{Runway} environment (\Cref{fig:mapping}).
BIEVR-LIO is the single geometry-only method that remains stable in all sequences and achieves the lowest error metrics on most. This demonstrates that our high-resolution mapping and informed sampling effectively leverages subtle geometric cues in geometrically uninformative environments. Notably, despite not using additional intensity information, BIEVR-LIO achieves more accurate odometry than COIN-LIO on all sequences except \textit{RunwayD}, in which the gyroscope is saturated for multiple consecutive frames, causing a rotational jump. Even then, the odometry remains consistent (RE).

\subsubsection{GEODE Dataset}
The GEODE~\cite{chen2024heterogeneous} dataset, summarized in \Cref{tab:geode}, comprises both handheld indoor sequences (\textit{Shield}, \textit{Tunneling}) and UGV-based outdoor sequences (\textit{Offroad}), all characterized by uninformative geometry captured on a Livox Avia LiDAR with an integrated IMU.
KISS-ICP degenerates on all sequences and is omitted from the comparison. The \textit{Tunneling} sequences, recorded in a mine, contain some informative structure from uneven tunnel walls and infrastructure, which most methods can exploit to achieve accurate odometry on \textit{Tunneling1}. \textit{Tunneling2} includes a short segment where the LiDAR faces directly towards the tunnel wall, causing several approaches to drift or diverge. BIEVR-LIO is able to use high-resolution irregularities of the stone surface and maintains accurate odometry, achieving the lowest ATE.
\textit{Offroad} sequences feature an open grass-field with limited geometric structure and pronounced sensor motion, resulting in poor performance for most methods as only FAST-LIO2 and BIEVR-LIO achieve accurate ATE in both sequences. The most challenging environment in our evaluation is \textit{Shield}, which consists of a straight metro tunnel. Here, longitudinal constraints only arise from small-scale structures such as the train bed and cutouts in the walls. These features are too subtle for most methods to exploit. In contrast, BIEVR-LIO captures these fine-grained geometric details and maintains robust odometry over the \SI{700}{\meter} \textit{Shield1} sequence (\Cref{fig:mapping}), even under fast rotational motion (\textit{Shield4}) or when the sensor directly faces the tunnel wall (\textit{Shield5}). 

\subsubsection{MARS-LVIG Dataset}
The MARS-LVIG dataset~\cite{li2024mars} consists of high-altitude UAV flights captured using a downward-facing Livox Avia with an integrated IMU. The sequences involve high velocities of up to \SI{6}{m/s} and cover distances of up to \SI{4.9}{\kilo\meter}. In \Cref{tab:mars}, we evaluate FAST-LIO2, iG-LIO, and BIEVR-LIO, as other methods failed during the take-off phase, where the LiDAR observes too few points for reliable state estimation.
Despite its high-resolution map, BIEVR-LIO robustly handles these sequences, in which the LiDAR point clouds are sparse due to flight altitudes of approximately \SI{80}{m}. It achieves performance comparable to the baselines on the \textit{AMtown02} and \textit{HKislandGNSS02} sequences. In the \textit{FeaturelessGNSS02} sequence, the UAV flies at low altitude over flat terrain, causing both iG-LIO and FAST-LIO2 to diverge. Owing to its fine-grained representation, BIEVR-LIO is still able to maintain stable odometry.

\begin{table}[b!]
\vspace{-2.5mm}
\caption{MARS-LVIG Dataset Results}
\label{tab:mars}
\vspace{-2.5mm}
\begin{adjustbox}{max width=\linewidth}
\begin{tabular}{lccc}
\toprule
\multicolumn{4}{c}{Absolute Trajectory Error RMSE (\SI{}{\meter}) / Relative Error RMSE (\textit{\%})} \\
\midrule
Method & AMtown02 & HKislandGNSS02 & FeaturelessGNSS02 \\
\midrule
FAST-LIO2 & $\underline{3.471}$ / $\underline{1.3}$ & $0.884$ / $1.4$ & $\times$ / $32.1$ \\
iG-LIO & $\mathbf{2.155}$ / $1.7$ & $\mathbf{0.2}$ / $\mathbf{0.4}$ & $\times$ / $\underline{25.6}$ \\
BIEVR-LIO & $4.0$ / $\mathbf{1.2}$ & $\underline{0.809}$ / $\underline{0.4}$ & $\mathbf{4.93}$ / $\mathbf{12.9}$ \\
\bottomrule
\end{tabular}
\end{adjustbox}
\vspace{-1.5mm}
\end{table}

\subsubsection{GrandTour Dataset}
Finally, we evaluate the GrandTour~\cite{Frey-Tuna-Fu-RSS-25} dataset, which features a quadruped platform in indoor and outdoor environments. The data exhibits slow yet shaky motion from legged locomotion and is captured using a Hesai XT32 LiDAR paired with a Honeywell HG4930 IMU. As this experiment is intended solely to demonstrate generalization to legged platforms and lower resolution LiDAR, we restrict the evaluation in \Cref{tab:grandtour} to LiDAR–inertial approaches.
On the \textit{LEICA-2} and \textit{TRIM-1} sequences, all methods achieve accurate odometry. In contrast, the confined sections present in \textit{ARC-4} and \textit{LEE-1} lead to drift or divergence for FAST-LIO2 and iG-LIO. BIEVR-LIO, however, remains accurate by exploiting fine-grained details, resulting in the most consistent performance across the  sequences.
\begin{table}[bt!]
\caption{GrandTour Dataset Results}
\label{tab:grandtour}
\begin{adjustbox}{max width=\linewidth}
\begin{tabular}{lcccc}
\toprule
\multicolumn{5}{c}{Absolute Trajectory Error RMSE (\SI{}{\meter}) / Relative Error RMSE (\textit{\%})} \\
\midrule
Method & ARC-4 & LEE-1 & LEICA-2 & TRIM-1 \\
\midrule
FAST-LIO2 & $0.872$ / $2.6$ & $2.949$ / $7.7$ & $\underline{0.027}$ / $0.3$ & $\mathbf{0.019}$ / $\mathbf{0.1}$ \\
DLIO & $\underline{0.052}$ / $0.6$ & $0.059$ / $0.4$ & $0.058$ / $0.7$ & $0.053$ / $0.3$ \\
iG-LIO & $\times$ / $32.1$ & $\times$ / $93.8$ & $0.051$ / $0.3$ & $\underline{0.023}$ / $\underline{0.2}$ \\
RESPLE & $0.067$ / $\mathbf{0.5}$ & $0.107$ / $0.4$ & $0.049$ / $0.4$ & $0.038$ / $0.2$ \\
RKO-LIO & $0.582$ / $1.1$ & $\underline{0.048}$ / $\underline{0.3}$ & $0.032$ / $\underline{0.2}$ & $0.031$ / $0.2$ \\
BIEVR-LIO & $\mathbf{0.051}$ / $\underline{0.6}$ & $\mathbf{0.024}$ / $\mathbf{0.1}$ & $\mathbf{0.015}$ / $\mathbf{0.1}$ & $0.028$ / $0.2$ \\
\bottomrule
\end{tabular}
\end{adjustbox}
\end{table}

\subsection{Ablation Study}
\label{sec:ablation}
In \Cref{tab:ablation_ate}, we present an ablation study to analyze the impact of our contributions. We perform representative experiments on the geometrically rich \textit{Cloister}, \textit{FieldS} with sparse structure, and the highly ambiguous \textit{Shield1} sequence containing only subtle geometric constraints. To isolate the effect of the proposed BIEVR-Map representation, we compare three configurations: first, using point-to-plane alignment against the principal voxel planes (Plane$|$$\times$$|$HR); second, also performing point-to-plane alignment, but using each BIEVR pixel as a high-resolution plane (Bump$|$$\times$$|$HR), without the additional Jacobian direction. In this case, the registration ignores neighboring pixels, effectively treating each pixel as an infinite plane. The third configuration (Plane$|$$\checkmark$$|$HR), enables the additional Jacobian direction, allowing the registration to fully exploit the surface variations captured by our map.
In the well-constrained \textit{Cloister} environment, all configurations perform similarly, as the largely planar structure is well represented by both map formulations. In \textit{FieldS}, the additional geometric detail provided by BIEVR leads to improved accuracy. In the most challenging \textit{Shield1} sequence, however, the benefits of BIEVR can only be exploited when the additional Jacobian direction is used, highlighting the importance of our registration formulation. 
We further evaluate the proposed map-informed dual-resolution sampling (ID) and compare it against fixed high-resolution sampling (HR) as well as a random dual-resolution strategy (RD) using the same resolutions (\SI{0.1}{\meter} and \SI{0.5}{\meter}), but selecting high-resolution voxels at random rather than based on map structure.
Across all sequences, RD performs worst, as high-resolution samples are often allocated to uninformative regions. Our proposed map-informed sampling consistently outperforms both baselines by focusing computation on informative areas, effectively improving the signal-to-noise ratio, while reducing the number of points used for registration by approximately a factor of four compared to HR sampling, resulting in lower runtime. A visual comparison of the runtime can be found in the appendix.

\begin{table}[t!]
\caption{Ablation Results}
\begin{adjustbox}{max width=\linewidth}
\begin{tabular}{lcc|ccc}
\toprule
Map & $\partial \mathbf{I}/\partial \boldsymbol{\xi}$ & Sampling & Shield1 & FieldS & Cloister \\
\midrule
& & &\multicolumn{3}{c}{Absolute Trajectory Error RMSE (\SI{}{\meter})} \\
\midrule
Plane & $\times$ & HR & $\times$ & $0.192$ & $\mathbf{0.051}$ \\
BIEVR & $\times$ & HR & $\times$ & $0.17$ & $0.058$ \\
BIEVR & $\checkmark$ & HR & $0.401$ & $\underline{0.163}$ & $0.059$ \\
BIEVR & $\checkmark$ & RD & $0.604$ & $0.218$ & $0.113$ \\
BIEVR & $\checkmark$ & ID & $\mathbf{0.256}$ & $\mathbf{0.159}$ & $\underline{0.053}$ \\
\midrule
& & &\multicolumn{3}{c}{Runtime per Frame (\si{\milli\second}) / Registration Points (\#)} \\
\midrule
BIEVR & $\checkmark$ & HR & $15.7$ / $13907$ & $17.2$ / $20423$ & $29.8$ / $33320$ \\
BIEVR & $\checkmark$ & ID & $12.8$ / $5039$ & $12.3$ / $5187$ & $22.6$ / $7574$ \\
\bottomrule
\end{tabular}
\end{adjustbox}
\vspace{-2mm}
\label{tab:ablation_ate}
\end{table}

\subsection{Runtime Analysis}
\begin{figure}[b]
\vspace{-4mm}
\includegraphics[width=\linewidth]{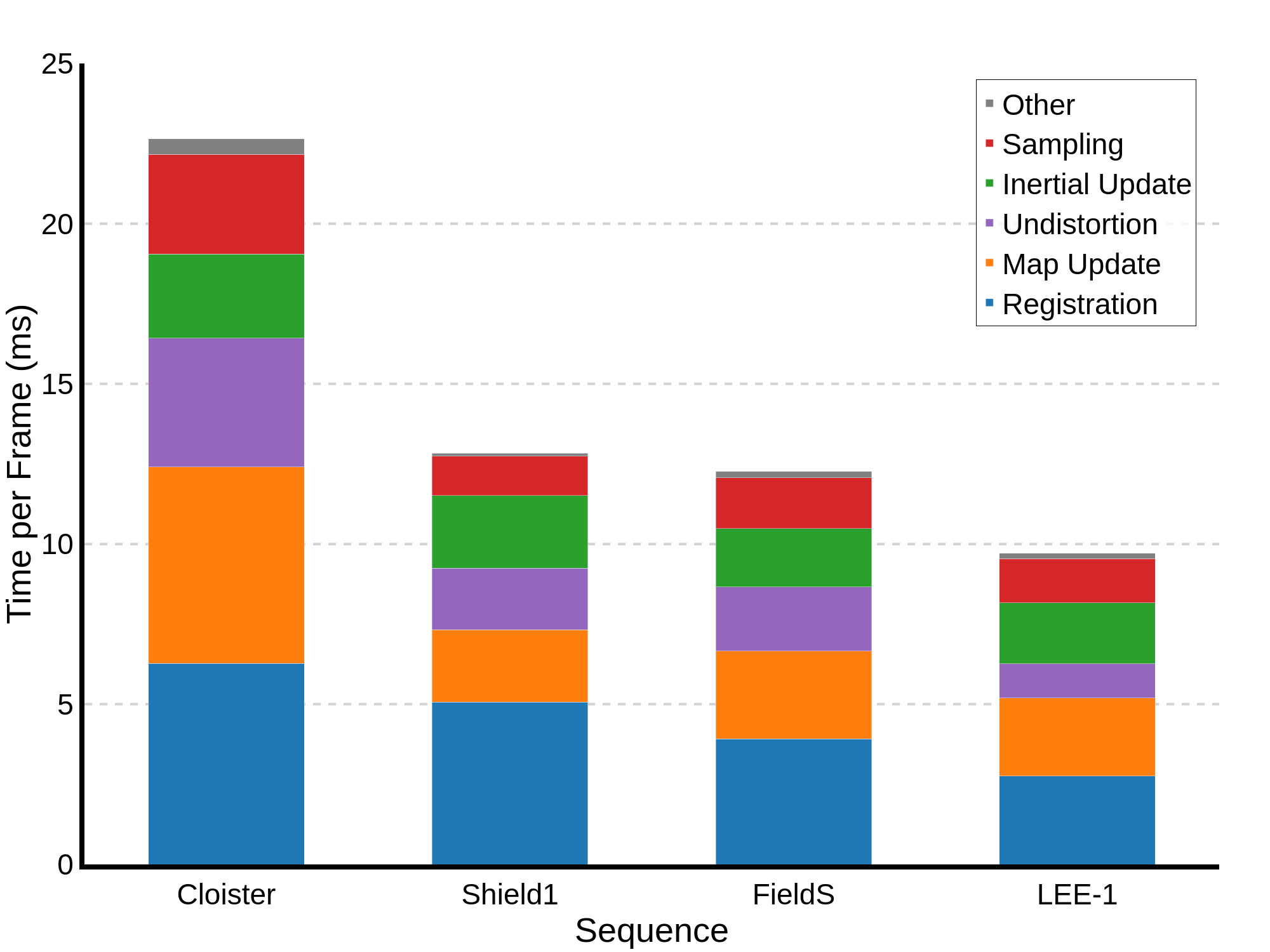}
\vspace{-7.5mm}
\caption{Runtime breakdown of BIEVR-LIO.}
\label{fig:runtime}
\end{figure}

As the runtime depends on the resolution of the LiDAR, as well as the surrounding environment, we report the average runtime per frame for four representative sequences. The \textit{Cloister} and \textit{FieldS} sequences were recorded using an Ouster OS-0 128-beam LiDAR, Shield1 was captured with a Livox Avia, and LEE-1 was recorded using a Hesai XT32.
The highest runtime is observed for the \textit{Cloister} sequence, which features large geometry due to tall surrounding buildings and is recorded with a high-resolution LiDAR. Despite being recorded with the same sensor, the runtime in \textit{FieldS} is roughly half of that, caused by very few LiDAR returns in the direction of the sky in the mostly flat environment. The lowest runtime is achieved on \textit{LEE-1}, which uses a lower-resolution 32-beam LiDAR.
As shown in \Cref{fig:runtime}, most of the computation time across all sequences is spent in the combined state update, which consists of the inertial update and the registration step. The time spent on map updates and sampling scales with the number of observed voxels and the number of input points. Notably, the map update, and thus also the undistortion, is performed on all input points. The inertial update runtime remains approximately constant, as it depends only on the window length, which is fixed to \SI{10}{\second} throughout our experiments.

\begin{figure}[t]
\includegraphics[width=\linewidth]{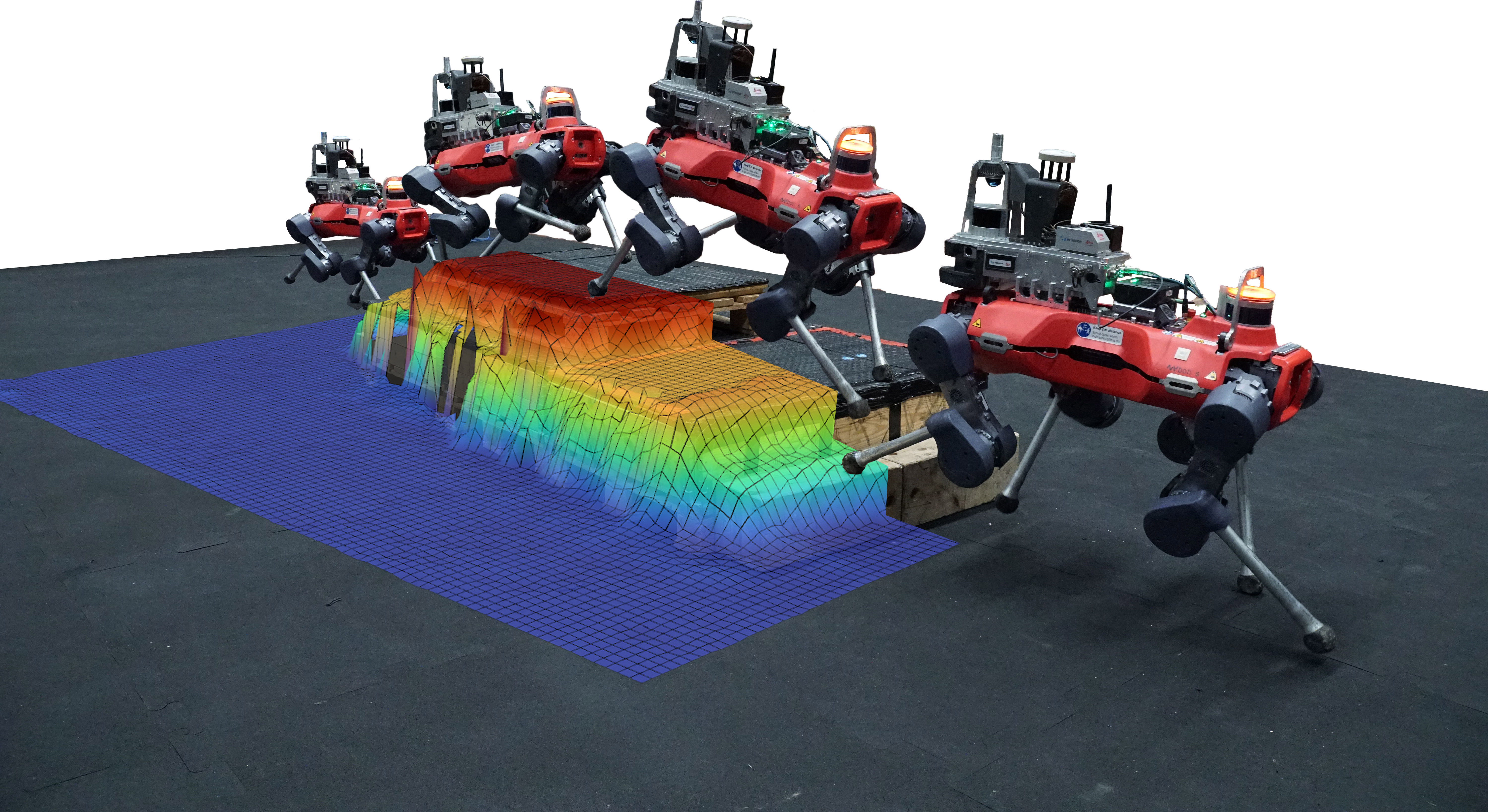}
\vspace{-7.5mm}
\caption{Locomotion planning based on BIEVR-LIO. The robot successfully traverses the obstacle. The extracted elevation map is partially visualized.}
\label{fig:traversal}
\vspace{-6mm}
\end{figure}

\subsection{Elevation Mapping}
\vspace{-0.5mm}
\label{sec:elevation}
We demonstrate the practicality of BIEVR-LIO for downstream applications in real-world locomotion experiments with a quadruped robot. The robot’s locomotion policy usually relies on an elevation map generated from depth cameras to plan its footsteps. In our experiments, we instead use a Livox Mid-360 LiDAR mounted at the front of the robot. We run BIEVR-LIO on the robot’s onboard computer and export the map in the format expected by the locomotion policy within a 4$\times$4 \SI{}{\meter} area around the robot. The generation and export of this elevation map takes approximately \SI{1}{\milli\second}.
In the first experiment, visualized in \Cref{fig:traversal}, the robot traverses an obstacle of stacked pallets. The proposed map representation provides sufficient detail for the robot to correctly place its feet on the surfaces of varying height. In the second experiment, shown in the supplementary video, the robot climbs a staircase, accurately placing its feet on each step. These results indicate that BIEVR-LIO captures sufficient geometric detail for locomotion planning and can serve as an alternative to separately constructed elevation maps, requiring only minimal additional computation beyond the odometry pipeline.
\vspace{-0.5mm}

\section{Limitations} 
\label{sec:limitations}
While the fine-grained resolution of BIEVR-LIO enables the use of subtle geometric details for registration, it also introduces some limitations. Sufficient point density is required to populate pixels, as no infilling is applied and empty pixels cannot be used, which may reduce performance for very low-resolution LiDARs or at high sensor velocities. Additionally, a reasonably accurate initial pose is needed to compute meaningful pixel residuals. Both issues could be mitigated using multi-scale image pyramids.

Our approach also does not exploit LiDAR intensity, making it less effective in strictly degenerate environments where geometry alone is insufficient, such as the \textit{Tunnel} scenario in the ENWIDE dataset. Intensity-based methods like COIN-LIO can still operate reliably in these cases.

While our approach reduces degeneracy, it does not explicitly detect it. As such, it could be naturally enhanced with existing degeneracy mitigation strategies as in \cite{huang2024lio}.

Finally, the system does not detect or model dynamic objects, which limits performance for downstream tasks, as the elevation mapping presented in \Cref{sec:elevation}.

\section{Conclusion} 
\label{sec:conclusion}
In this work, we presented BIEVR-LIO, a novel LIO framework built around a high-resolution yet efficient map representation that captures subtle geometric details using voxel-based height images. The proposed pipeline further introduces a map-informed point sampling strategy that focuses computation on informative regions, improving robustness while reducing runtime compared to a global high-resolution sampling. Experiments across multiple sensors, platforms, and environments demonstrate that our approach significantly improves performance in geometrically challenging scenarios, enabling accurate registration where existing methods fail and achieving state-of-the-art results in well-structured environments. Notably, all experiments were conducted using the same parameter set, highlighting the robustness and generalizability of the proposed method across diverse conditions.
In addition, we show that the fine-grained geometry encoded in BIEVR-Map is well suited for elevation mapping, while remaining computationally efficient enough to operate in real-time. In future work, we plan to incorporate LiDAR intensity information to further extend the range of environments in which BIEVR-LIO can operate reliably.


\bibliographystyle{unsrtnat}
\bibliography{references}

\clearpage
\section*{APPENDIX}
\subsection{Parameter Analysis}
We provide experimental results on the influence of bump-image resolution $r$ and voxel length $l$, for the proposed informed sampling (ID) and global high-resolution sampling (HR) at \SI{0.1}{\meter} in \Cref{tab:ablation_ate}. Note that $l=\SI{1}{\meter}$ tracks coarser planes than $l=\SI{0.5}{\meter}$, but adds more points to the registration under ID. In the structure-rich \textit{Cloister}, results are almost identical in all settings. In the more challenging \textit{FieldS}, all settings achieve accurate odometry, with minor improvements at smaller $r$. In the very challenging \textit{Shield1}, informative geometry is limited to fine-grained details. $r=\SI{0.15}{\meter}$ fails to capture them and diverges, \SI{0.1}{\meter} remains stable but with larger drift, and \SI{0.05}{\meter} achieves the best results. This supports our hypothesis that high resolution is critical to extract constraints from fine-grained structure. The experiments show that our method is reasonably robust to parameter changes.
\begin{table}[h!]
\caption{Parameter Analysis: Absolute Trajectory Error RMSE (\SI{}{\meter})}
\vspace{-2mm}
\begin{adjustbox}{max width=\linewidth}
\begin{tabular}{lcc|ccc}
\toprule
$r$ & $l$ & Sampling & Shield1 & FieldS & Cloister \\
\midrule
0.05 & 0.5 & ID / HR & $\underline{0.256}$ / $0.401$ & $\mathbf{0.159}$ / $0.163$ & $0.053$ / $0.059$ \\
0.05 & 1.0 & ID / HR & $\mathbf{0.232}$ / $0.354$ & $0.168$ / $0.163$ & $0.055$ / $\underline{0.048}$ \\
0.10 & 0.5 & ID / HR & $\times$ / $0.893$ & $\underline{0.161}$ / $0.166$ & $0.057$ / $0.057$\\
0.10 & 1.0 & ID / HR & $0.409$ / $0.815$ & $0.164$ / $0.168$ & $0.064$ / $\mathbf{0.047}$\\
0.15 & 0.5 & ID / HR& $\times$ / $\times$ & $0.172$ /  $0.167$ & $0.066$ / $0.053$ \\
0.15 & 1.0 & ID / HR & $\times$ / $\times$ & $0.165$ /  $0.174$ & $0.092$ / $0.048$ \\
\bottomrule
\end{tabular}
\end{adjustbox}
\label{tab:ablation_ate}
\vspace{-4mm}
\end{table}

\subsection{Runtime Ablation}

\begin{figure}[b]
\vspace{-5mm}
\includegraphics[width=\linewidth]{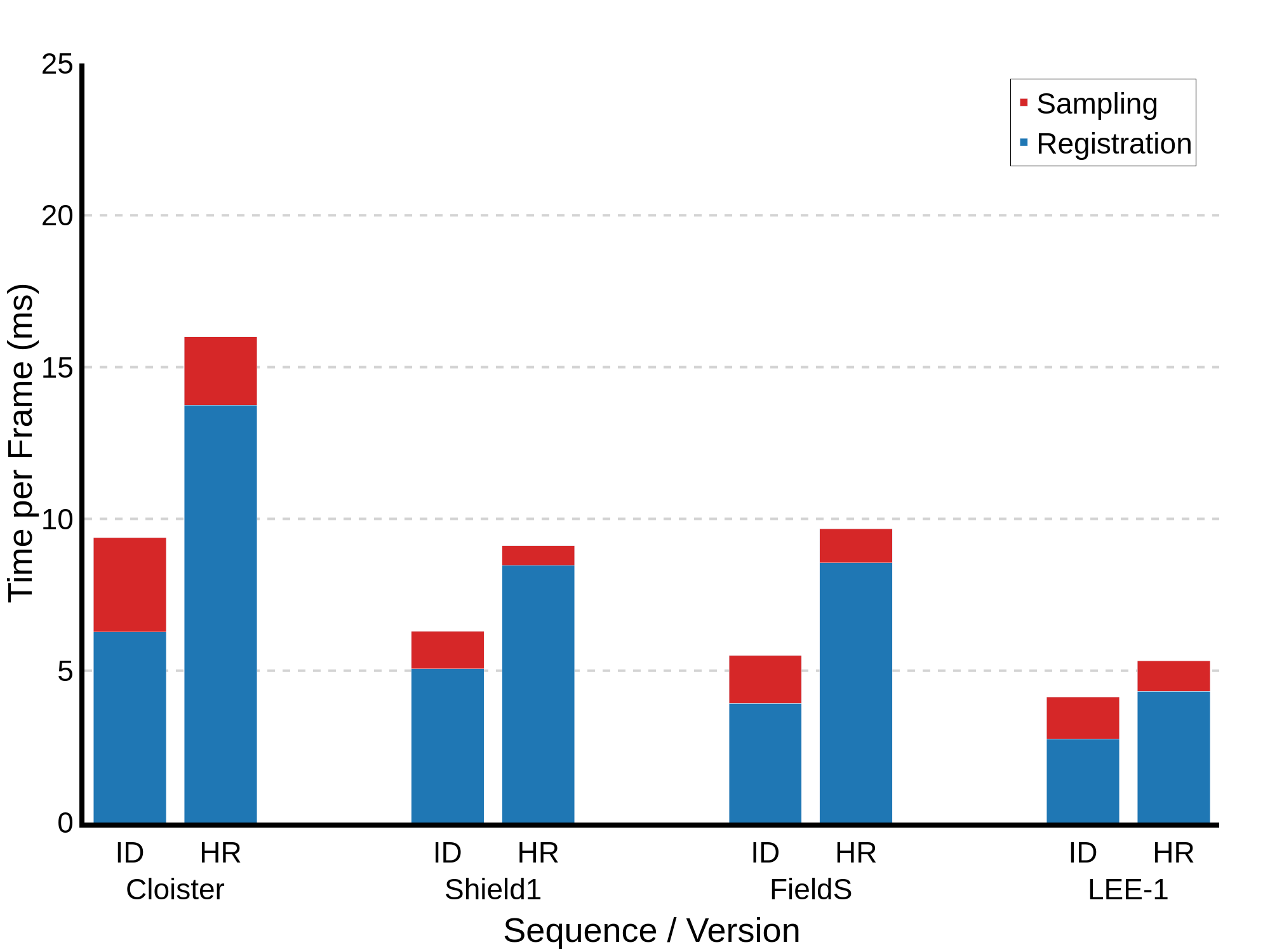}
\vspace{-7.5mm}
\caption{Runtime comparison between the proposed Informed Dual-Resolution (ID) and High-Resolution (HR) sampling.}
\label{fig:runtime_sampling}
\end{figure}

We compare the runtime of our informed sampling (ID) against a global high-resolution sampling (HR) in \Cref{fig:runtime_sampling}. The runtime of the registration is significantly reduced by our map-informed sampling strategy. Although the two-stage sampling strategy requires slightly more time than simple high-resolution sampling, the reduced number of points used for registration leads to a lower overall runtime. This improvement is most pronounced in the \textit{Cloister} sequence, which contains high-resolution point clouds. In contrast, the effect is only marginal in the \textit{LEE-1} sequence, as the low-resolution point clouds rarely contain multiple points per voxel.

\subsection{Registration Residual Jacobian}
The residual Jacobian consists of two terms:
\begin{equation}
\frac{\partial r}{\partial \boldsymbol{\xi}} =  
\underbrace{\frac{\partial [{}_\mathrm{C}\mathbf{p}]_z}{\partial \boldsymbol{\xi}}}_{\textit{I}} - \underbrace{\frac{\partial \mathbf{I}(u, v)}{\partial \boldsymbol{\xi}}}_{\textit{II}} 
\label{eq:jacobian}
\end{equation}

The first term (\textit{I}) describes the Jacobian of the residual component of the point height:
\begin{equation}
\frac{\partial [{}_\mathrm{C}\mathbf{p}]_z}{\partial \boldsymbol{\xi}} = \frac{\partial [{}_\mathrm{C}\mathbf{p}]_z}{\partial {}_\mathrm{C}\mathbf{p}} \cdot \frac{\partial {}_\mathrm{C}\mathbf{p}}{\partial \boldsymbol{\xi}}
\end{equation}

The second term (\textit{II}) describes the Jacobian of the residual component of the height image pixel value: 
\begin{equation}
\frac{\partial \mathbf{I}(u, v)}{\partial \boldsymbol{\xi}} =  \frac{\partial\mathbf{I}(u, v)}{\partial (u,v)} \cdot \frac{\partial (u,v)}{\partial [{}_\mathrm{C}\mathbf{p}]_{xy}} \cdot \frac{\partial[{}_\mathrm{C}\mathbf{p}]_{xy}}{\partial {}_\mathrm{C}\mathbf{p}}  \cdot \frac{\partial {}_\mathrm{C}\mathbf{p}}{\partial \boldsymbol{\xi}}
\end{equation}

The derivatives of the individual point components are:
\begin{equation}
\frac{\partial[{}_\mathrm{C}\mathbf{p}]_{xy}}{\partial {}_\mathrm{C}\mathbf{p}} = 
\begin{bmatrix}
1 & 0 & 0 \\
0 & 1 & 0
\end{bmatrix} 
\quad \frac{\partial [{}_\mathrm{C}\mathbf{p}]_z}{\partial {}_\mathrm{C}\mathbf{p}} = [0\quad 0\quad 1]
\end{equation}

The derivatives of the pixel position are defined as:
\begin{equation}
\frac{\partial (u,v)}{\partial [{}_\mathrm{C}\mathbf{p}]_{xy}} = \begin{bmatrix}
r^{-1} & 0 \\
0 & r^{-1}
\end{bmatrix}
\end{equation}
and $\frac{\partial\mathbf{I}(u, v)}{\partial (u,v)}$ is the image gradient from neighboring pixels.

Finally, as stated in \cite{sola2018micro}, the $SE(3)$ Jacobian of the rigid motion action is:
\begin{equation}
\frac{\partial {}_\mathrm{C}\mathbf{p}}{\partial \boldsymbol{\xi}} = \Big[\mathbf{R}_\mathrm{CI}^* \quad -\mathbf{R}_\mathrm{CI}^*[{}_\mathrm{I}\mathbf{p}]_\times \Big]
\end{equation} 
where $\mathbf{R}_\mathrm{CI}^*$ is the rotational component of $\mathbf{T}_\mathrm{CI}^* = \mathbf{T}_\mathrm{CG} \, \mathbf{T}_\mathrm{GI} \, \mathrm{Exp(\boldsymbol{\xi})}$ and $[\mathbf{p}]_\times$ indicates the skew-symmetric matrix of a point.

\end{document}